\documentclass{article}

\usepackage{microtype}
\usepackage{graphicx}
\usepackage{subfigure}
\usepackage{booktabs}
\usepackage[table]{xcolor}
\usepackage{hyperref}
\usepackage{amsmath}
\usepackage{amssymb}

\usepackage[accepted]{mlsys2025}
\makeatletter
\renewcommand{\Notice@String}{}
\makeatother

\newcommand{\hd}[2]{\begin{tabular}[b]{@{}c@{}}#1\\#2\end{tabular}}  %
\definecolor{hlrow}{gray}{0.92}
\newcommand{\nb}{\rowcolor{hlrow}}  %

\mlsystitlerunning{NoeVera: Can Knowledge Guide Experiments and Experiments Refine Knowledge?}

\begin{document}

\twocolumn[

\mlsystitle{NoeVera: Can Knowledge Guide Experiments and Experiments Refine Knowledge?}

\mlsyssetsymbol{equal}{*}

\begin{mlsysauthorlist}
\mlsysauthor{Pinaki Prasad Guha Neogi}{ms,to}
\mlsysauthor{Spyros Mastorakis}{ms}
\mlsysauthor{Ankur Mallick}{ms}
\mlsysauthor{Saravan Rajmohan}{ms}
\mlsysauthor{Victor Rühle}{ms}
\end{mlsysauthorlist}

\mlsysaffiliation{ms}{Microsoft M365 Copilot Research}
\mlsysaffiliation{to}{The Ohio State University (Work done during internship at Microsoft)}

\mlsyscorrespondingauthor{Pinaki Prasad Guha Neogi}{guhaneogi.2@osu.edu}

\mlsyskeywords{LLM serving, research agents, retrieval-augmented generation, performance simulation, knowledge management}

\vskip 0.3in

\begin{abstract}
\vspace{2mm}
With the growing demand for large language model (LLM) services, optimizing LLM serving systems is becoming increasingly complex, and the bottleneck is no longer predicting how a given configuration will perform. The harder problem is deciding what to investigate: which mechanisms matter for a workload, which configurations are worth testing, and what evidence a decision still needs.
In this work, we present \textsc{NoeVera}, an agentic \emph{Second Brain} that goes beyond just one-shot retrieval by turning papers, code, benchmarks, and experimental results into a persistent, provenance-preserving knowledge base with three planes: source evidence, reusable synthesis, and agent memory.
Decision-aware retrieval assembles compact evidence packs, role-specific agents reason over them and invoke a simulator (Vidur or GAUSS) whenever quantitative support is needed, and every finding is written back as a scoped evidence record, so knowledge guides experiments and experiments refine knowledge.
In configuration search over four serving scenarios, NoeVera comes within 5\% of the optimum in a median of 6--7 simulator calls, versus 9 for Bayesian optimization and 15 for a knowledge-free agent with the same tools; on A100 GPUs, its representative configurations deliver $\sim1.30\times$ vLLM-default throughput and meet every scenario's latency objective.
When reused, provenance-preserving evidence records over-generalize on only 8\% of applicability traps, versus 28\% for a general-purpose agent memory.
On the held-out 48-task split of a 72-task benchmark, NoeVera scores 29\% higher than an equally equipped agent and 36\% higher than its single-call retrieval counterpart (0.80 vs.\ 0.62 and 0.59), and even when baselines receive NoeVera's own prompts, a statistically significant part of its evidence-evolution advantage remains.
\end{abstract}
]

\printAffiliationsAndNotice{}

\section{Introduction}
\label{sec:intro}

\begin{figure*}[t]
\centering
\includegraphics[width=\linewidth]{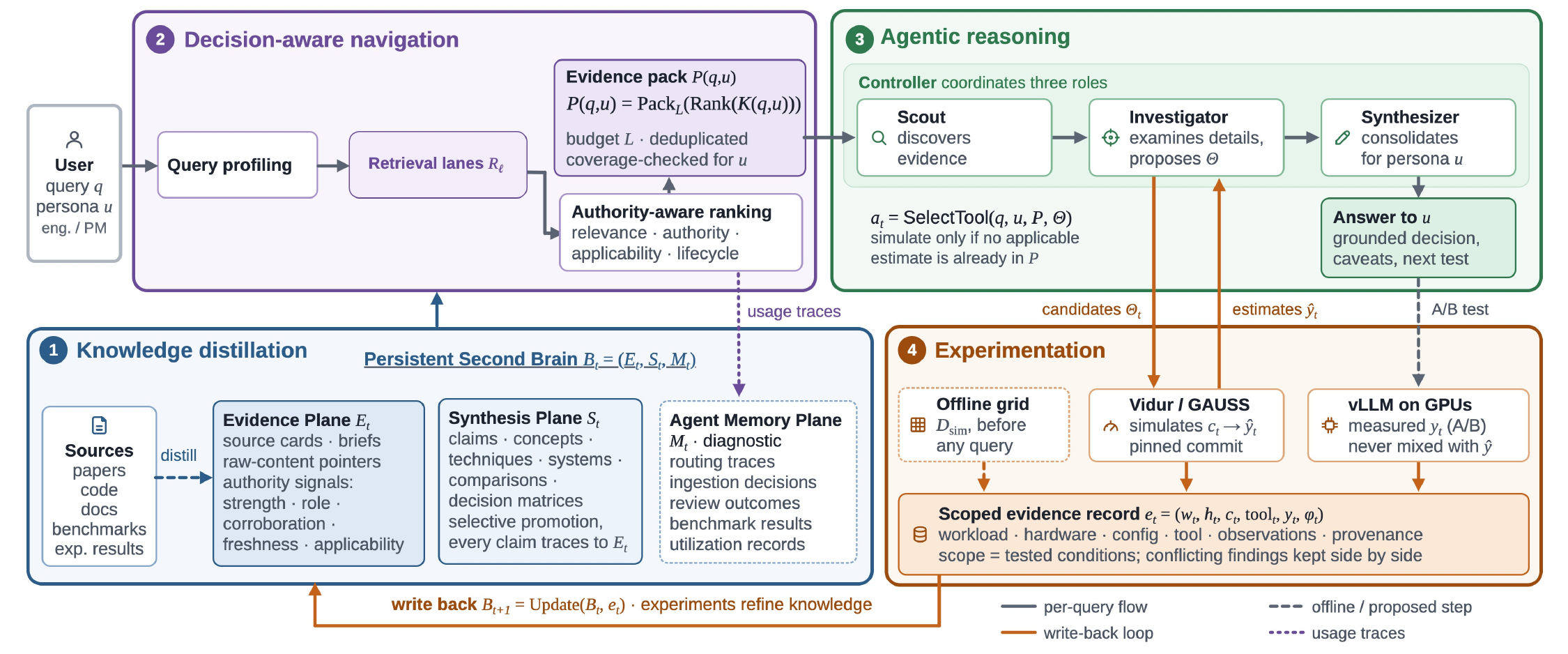}
\caption{NoeVera architecture. Sources are distilled into three planes of a persistent Second Brain; for a query $q$ and persona $u$, the hybrid retrieval lanes feed authority-aware ranking and a budgeted evidence pack; scout, investigator, and synthesizer roles reason over the pack and, when needed, run simulator experiments (Vidur by default; GAUSS in \S\ref{sec:res_c1}), whose results return as scoped evidence records.}
\label{fig:arch}
\end{figure*}

Consider a researcher who must raise the throughput of Llama-3-70B on a four-GPU A100 node without letting the 95th-percentile time-to-first-token (TTFT) exceed 400~ms (a relaxed variant of the Sc1 scenario in \S\ref{sec:res_c1}, which uses a 300~ms limit).
Should they enable chunked prefill, turn on prefix caching, or change batching limits?
The answer depends on the model, hardware, and workload, but also on how these mechanisms interact, what prior work has established, and which experiments could tell the candidates apart.
For workload $w$ and hardware $h$, the researcher seeks
\begin{equation}
\label{eq:objective}
\begin{aligned}
\max_{c\in\mathcal{C}}\ \ & \mathrm{Capacity}(c;w,h)\\
\text{s.t.}\ \ & \mathrm{TTFT}_{p95}(c;w,h)\le\tau,\quad \mathrm{Mem}(c;w,h)\le M_h,
\end{aligned}
\end{equation}

where capacity is the maximum sustainable generated-token rate of serving configuration $c$, $\tau$ is the latency service-level objective (SLO), and $M_h$ is per-device memory budget.
Throughout, \emph{throughput} denotes this capacity, not the offered arrival rate. We determine it by an arrival-rate sweep to identify the highest rate at which the queue remains stable, \& evaluate TTFT$_{p95}$ \& memory at the scenario's offered load, a fixed operating point.
A simulator such as Vidur~\cite{agrawal2024vidur} or GAUSS~\cite{bharadwaj2026gauss} can easily estimate any configuration it is given, but the harder problem is constructing a meaningful search space $\mathcal{C}$ in the first place: deciding \emph{which configurations are worth exploring, why they might work, and what evidence is still missing}.

\vspace{2mm}

Existing tools each address part of this problem.
Serving frameworks such as vLLM~\cite{kwon2023efficient} expose a growing set of interacting mechanisms; simulators explore configuration spaces without repeated GPU runs; retrieval-augmented generation (RAG) supplies external information; and agentic systems such as Glia~\cite{hamadanian2025glia} show that language models can take part in systems design through iterative experimentation.
Yet the knowledge connecting these capabilities is scattered across papers, code, documentation, benchmarks, and past experiments, each with its own assumptions and reliability.
A retrieved passage can name the right mechanism while describing an incompatible workload; a simulator estimate does not say which hypothesis motivated it or how it should shape the next experiment; and without persistent memory, agents rebuild context for every question and lose what earlier investigations found.

We present \textsc{NoeVera}, an agentic \emph{Second Brain} that turns heterogeneous sources into a persistent, evidence-grounded substrate for LLM systems research.
Its \emph{Evidence Plane} preserves sources and their provenance, its \emph{Synthesis Plane} selectively distills reusable claims, mechanisms, and decision matrices, and its \emph{Agent Memory Plane} records how knowledge is used.
Decision-aware retrieval ranks candidates by authority and applicability as well as relevance and assembles compact evidence packs; role-specific agents reason over them, decide when a question needs quantitative support, and propose which configurations to simulate; and simulator results, from an offline grid and on-demand runs, return as scoped evidence records, so knowledge guides experiments and experiments refine knowledge.

We evaluate NoeVera through controlled LLM-serving experiments
(\S\ref{sec:results}), examining three central claims.
First, knowledge-guided search reaches near-optimal configurations
with fewer simulator calls than Bayesian optimization or a
knowledge-free agent, using either Vidur or GAUSS. The resulting
configurations deliver 1.30--1.31$\times$ vLLM-default throughput
on real hardware (C1).
Second, scoped evidence records enable safe reuse of experimental
findings, with 8\% over-generalization versus 28\% for agent memory
and a 64\% reduction in simulator calls per query over a query
stream (C2).
Third, structured knowledge and orchestration improve research
decisions, particularly their revision when evidence changes,
beyond what prompt matching alone achieves (C3).
To support these capabilities, we make the following contributions:

\begin{itemize}
    \item \textbf{A provenance-preserving Second Brain for research decisions:}
    a three-plane knowledge architecture that separates source
    evidence, supported synthesis, and agent-use memory, enabling
    traceable reasoning and preserving the conditions under which
    technical findings remain applicable
    (\S\ref{sec:brain}).

    \item \textbf{Decision-aware reasoning and knowledge-guided
    experimentation:}
    an agentic workflow that combines authority- and
    applicability-aware evidence navigation with selective
    simulation, using accumulated knowledge to identify
    promising configurations and determine when new
    quantitative evidence is needed
    (\S\ref{sec:nav}, \S\ref{sec:reason}).

    \item \textbf{Closed-loop evidence reuse and systematic evaluation:}
    a workflow that writes experimental findings back with
    explicit applicability conditions, enabling reuse across
    questions without treating prior results as universally
    valid; %
    evaluated in configuration search with Vidur and GAUSS and on NoeVeraBench- a 72-task benchmark with 24 development and 48 held-out
    tasks
    (\S\ref{sec:loop}, \S\ref{sec:results},
    \S\ref{sec:method}).
\end{itemize}

\section{NoeVera Design}
\label{sec:design}

Figure~\ref{fig:arch} illustrates NoeVera's four stages: knowledge distillation into a persistent Second Brain, decision-aware navigation, agentic reasoning, and experimentation with results written back to the knowledge base. The design follows four goals: \textbf{G1}, preserve original evidence while enabling traceable synthesis (\S\ref{sec:brain}); \textbf{G2}, retrieve evidence for decisions rather than similarity alone (\S\ref{sec:nav}); \textbf{G3}, use knowledge to decide when and what to simulate (\S\ref{sec:reason}); and \textbf{G4}, make findings persistent and reusable while preserving conflicts and limiting conclusions to tested conditions (\S\ref{sec:loop}).

\subsection{An Evidence-Preserving Second Brain}
\label{sec:brain}

We represent the persistent knowledge state as $\mathcal{B}_t=(\mathcal{E}_t,\mathcal{S}_t,\mathcal{M}_t)$: source evidence, synthesized knowledge, and agent memory, which are logical planes rather than physically separate stores.
Distillation preserves source coverage, promotes knowledge selectively, and keeps every derived conclusion traceable to its evidence.
Our knowledge base represents 2,627 source files (papers, documentation, repositories, benchmark reports, and experimental results)%
, and organizes the derived knowledge
into 2,912 wiki pages, and indexes it as 87,357 typed
retrieval chunks.

\vspace{2mm}

\textbf{Evidence Plane.}
Each source has a source card, a source brief, and pointers back to its raw content, allowing reasoning to recover conditions that summaries may omit, such as the workload, batch size, or memory limit under which a throughput gain was measured. The plane also carries authority signals: evidence strength, source role, corroboration, freshness, and applicability. These signals are not interchangeable: a foundational paper may remain authoritative as it ages, while a recent measurement may be relevant but uncorroborated. Thus, freshness provides context rather than a rule that newer evidence takes precedence. Sources not yet suitable for synthesis remain available as anchor or deferred evidence than being forced into general claims.

\vspace{2mm}

\textbf{Synthesis Plane.}
Source observations are distilled into claims, concepts, techniques, systems, clusters, comparisons, and decision matrices that connect otherwise separate sources, for instance observations about memory pressure and latency spread across prefix-caching, KV-cache, and scheduling papers.
Synthesis is selective: only evidence that yields reusable understanding is promoted. Every derived object $s$ retains a support set $\mathrm{Support}(s)\subseteq\mathcal{E}_t$, non-empty for an evidence-backed claim, so decision matrices remain traceable syntheses rather than independently verified facts.

\vspace{2mm}

\textbf{Agent Memory Plane.}
Routing traces, ingestion decisions, review outcomes, benchmark results, and knowledge-utilization records capture how the system uses its knowledge.
They show why an answer is unsupported (a missing source, failed retrieval, a dropped condition, or faulty reasoning over adequate evidence) and thus where a fix belongs.
This diagnostic memory is distinct from experimental findings, which enter as domain evidence (\S\ref{sec:loop}), and in every reported experiment it is excluded from retrieval at inference time: it is used for analysis, not for answering.

\vspace{1mm}

\subsection{Decision-Aware Navigation}
\label{sec:nav}

Textual relevance alone does not guarantee sufficient evidence:
a passage may use the right terminology while describing a
different workload or a superseded implementation.
NoeVera addresses this by combining complementary retrieval
signals with authority- and applicability-aware ranking,
then assembling a compact evidence pack tailored to the
decision at hand.

\textbf{Query profiling.}
The profile records query terms, exact technical phrases, configuration identifiers, and the intent: concept explanation, troubleshooting, method comparison, optimization decision, or quantitative what-if analysis.
It also records the persona $u$, because an engineer diagnosing a regression and a product manager evaluating a rollout need different evidence; the persona shapes retrieval priorities, not only presentation.

\textbf{Hybrid retrieval across complementary signals.}
The candidate set $\mathcal{K}(q,u)$ is the union of results from a deliberately chosen set of complementary retrieval mechanisms: lexical BM25 and exact technical matching, dense semantic retrieval, knowledge-graph expansion and navigation-hub retrieval, code-oriented retrieval, and fine-grained evidence-atom retrieval. The design principle is coverage, not count: exact configuration names and API identifiers require precise lexical matching, conceptual questions require semantic similarity, cross-technique comparisons require graph and navigation access, code-level questions require repository-aware retrieval, and evidence-grounded reasoning requires atomic passages that survive packing. Combining these signals lets $\mathcal{K}(q,u)$ cover all of these query classes rather than biasing toward any one.

\vspace{1mm}

\textbf{Authority-aware ranking and evidence packs.}
Candidates are ranked by query relevance, metadata fit, persona compatibility, evidence strength, applicability, freshness, and lifecycle authority, combined as a normalized weighted sum whose weights are tuned on the development split and then fixed; material under review stays discoverable with reduced recommendation authority, and superseded material provides history rather than guidance.
NoeVera then packs $\mathcal{P}(q,u)=\mathrm{Pack}_{L}(\mathrm{Rank}(\mathcal{K}(q,u)))$ under a token budget $L$, balancing directly relevant evidence with supporting mechanisms, risks, and validation requirements while removing redundancy.
Each pack is checked against persona-specific coverage requirements (mechanisms, telemetry, interventions, risks, validation, and provenance for engineering questions; decisions, benefits, risks, metrics, rollout conditions, and provenance for product questions); coverage is a diagnostic signal, not proof of correctness.

\subsection{Agentic Reasoning and Knowledge-Guided Experimentation}
\label{sec:reason}

Building on the retrieved evidence pack, NoeVera uses a controller to coordinate three roles: a \emph{scout} that discovers evidence, an \emph{investigator} that examines technical details and open questions, and a \emph{synthesizer} that consolidates findings for the persona. Engineering responses emphasize mechanisms, workload assumptions, interventions, and validation steps; product (PM) responses emphasize benefits, trade-offs, risks, success metrics, and conditions for revisiting a decision. In both modes, the agents distinguish supported observations from inferred recommendations and preserve uncertainty rather than turning a plausible mechanism into an unsupported number.

\textbf{Deciding when and what to simulate.}
The controller decides $a_t=\mathrm{SelectTool}(q,u,\mathcal{P},\Theta)$: whether existing evidence suffices or a quantitative tool is needed, where $\Theta$ is a set of candidate experimental settings.
Crucially, $\Theta$ need not come from the user: the investigator extracts, from the pack's mechanisms and benchmark reports, parameter combinations reported as viable under the target workload's prompt-length and concurrency profile, and filters them against the hardware's memory budget.
The controller invokes a simulator when the query profile, built from the query text alone before retrieval, marks the decision as needing a quantitative estimate and the pack contains no directly applicable estimate for the target workload and hardware.
Two pathways supply quantitative evidence.
Before any query, a systematic offline grid $\mathcal{D}_{\mathrm{sim}}=\{(w_i,h_i,c_i,\hat{\mathbf{y}}_i)\}$ of simulator runs over relevant serving parameters enriches the knowledge base with estimates tied to their conditions.
At query time, the agent runs a simulator on selected candidates $c_t$ for the target $(w_t,h_t)$, in parallel with its ongoing investigation, and interprets the estimates $\hat{\mathbf{y}}_t$ alongside the retrieved mechanisms.
Simulated estimates and hardware measurements are never conflated: an estimate inherits the simulator's assumptions, whereas a measurement holds only for its runtime, workload, and hardware.
Rather than solving a global optimization problem,
NoeVera connects evidence-grounded reasoning to the
selection of informative experiments.
Vidur~\cite{agrawal2024vidur} is our primary on-demand backend for this work, but the simulator is swappable: \S\ref{sec:res_c1} repeats the search with GAUSS~\cite{bharadwaj2026gauss}, an analytical + Monte-Carlo simulator.

\subsection{Closing the Loop: Scoped, Reusable Evidence}
\label{sec:loop}

Every experimental finding is stored as an evidence record $e_t=(w_t,h_t,c_t,\mathrm{tool}_t,\mathbf{y}_t,\phi_t)$, holding the workload, hardware, configuration, tool or method (e.g., a pinned Vidur commit), quantitative observations, and provenance, and is written back: $\mathcal{B}_{t+1}=\mathrm{Update}(\mathcal{B}_t,e_t)$.
A record's scope is explicit: one simulation supports a statement about that configuration, not a general claim, and it is promoted into synthesis only after its scope and evidential role are assessed.
Findings that reinforce, qualify, or contradict earlier evidence are kept side by side with their conditions, so later reasoning sees a conflict instead of silently replacing one result with another.
When a later query retrieves a record, an applicability check compares its workload and hardware (model family and size, precision, GPU SKU, and prompt/output length distribution, within fixed tolerances) and recorded assumptions with the query: a full match returns the record as applicable, a partial match returns it as historical context with an explicit caveat, and a mismatch rejects it.
The loop $\mathcal{B}_t\rightarrow\Theta_t\rightarrow e_t\rightarrow\mathcal{B}_{t+1}$ thus lets knowledge select experiments whose results become evidence for later decisions.

\vspace{1mm}

\textbf{Example.}
Consider the Llama-3-70B optimization scenario introduced
in \S\ref{sec:intro}: BF16 with TP=4 on a
4$\times$A100-80GB node, 512/128-token requests at
8 requests/s, and TTFT$_{p95}\le400$~ms.
NoeVera profiles the query as a quantitative what-if with
$M_h=76$~GB and assembles a 14-item, 2,480-token
evidence pack (including two measured benchmarks).
Based on the retrieved evidence, it proposes three
configurations (chunk size, prefix caching, maximum
sequences): (512, on, 128), (1024, on, 96), and
(256, off, 192).
Three Vidur calls estimate capacities of 1,180, 1,290,
and 1,050 generated tokens/s, respectively, with
TTFT$_{p95}$ of 388, 402, and 362~ms at the offered
load of 1,024 tokens/s.
NoeVera recommends the first configuration, the
highest-capacity candidate satisfying the SLO.
It states the assumptions (the 512/128 workload holds
and the prefix-cache hit rate is at least 0.3),
proposes a vLLM A/B test, and writes back the finding. The value of this write-back becomes apparent in a
later query. At query 17 of the \S\ref{sec:res_c2}
stream, a user reports \emph{delivered} throughput
of 920 tokens/s at the same offered load and asks
whether the stored 1,180-token/s estimate indicates
unrealized headroom or an underperforming deployment.
NoeVera finds that the model, GPU, and workload
match, but the assumed cache hit rate remains
unverified. It therefore treats the finding as
partially applicable, surfaces the estimate with
this caveat, and requests the current hit rate,
which would save three Vidur calls once confirmed
(full trace in Appendix~E.7).

\section{Evaluation Methodology}
\label{sec:method}

We structure the evaluation around three claims,
holding task specifications and tool interfaces fixed
across comparisons while explicitly accounting for
differences in prior knowledge.
\textbf{C1, knowledge-guided experiment selection:}
structured knowledge reduces the new simulator evaluations
needed to find good configurations.
\textbf{C2, safe reuse:} written-back findings improve later
decisions without being over-generalized to mismatched
conditions.
\textbf{C3, evidence-grounded reasoning:} structured
knowledge and orchestration improve the correctness and
revision of research decisions.
C1's search metrics and C2's primary metric (the
over-generalization rate) do not depend on an LLM judge;
C3 is judge-scored, with its confirmatory test on the
evidence-evolution track.
We call results \emph{descriptive} when no hypothesis
test is reported (mainly the N track, and
Table~\ref{tab:e2e}'s E scores, which feed
Table~\ref{tab:evo}'s confirmatory test),
\emph{diagnostic} for the D track and the non-statistical
C1 analyses, \emph{confirmatory} only for the two tests
of the confirmatory family, and \emph{exploratory} for
other comparisons reported with confidence intervals
($\Delta_{\mathrm{reuse}}$, the query-stream gain,
D-track paired comparisons, and the robustness gaps
of \S\ref{sec:res_cost}).

\textbf{Systems.}
We compare against an LLM-only reference; Dense RAG
(dense top-$k$ retrieval over the raw files); Hybrid RAG;
GraphRAG~\cite{edge2024graphrag},
HippoRAG~2~\cite{gutierrez2025hipporag2}, and
LightRAG~\cite{guo2025lightrag} with their authors'
defaults; PaperQA2~\cite{skarlinski2024paperqa2},
an agentic literature-synthesis system evaluated on the
navigation track; and ReAct, a generic ReAct-style
agent~\cite{yao2023react} with hybrid search and a
document reader.
Hybrid RAG uses NoeVera's hybrid retrieval and
authority- and lifecycle-aware ranking over the raw files
and the Evidence Plane's source representations and
authority metadata, but excludes the Synthesis and
Agent Memory planes and operates in a single call.
It therefore shares NoeVera's retrieval substrate,
but neither its synthesized knowledge nor its orchestration.
The suffix +V adds Vidur through one shared tool interface
and its documentation: ReAct+V is the baseline most
comparable to NoeVera+V, OpenHands+V gives
OpenHands~\cite{wang2025openhands} the raw file tree,
\texttt{grep}, and the Vidur command-line tool, and
LLM-only+V and Hybrid RAG+V separate the simulator's
value from that of knowledge (configurations in
Appendix~A.2).
NoeVera+G uses GAUSS~\cite{bharadwaj2026gauss} as the
on-demand backend instead (\S\ref{sec:res_c1} only).
We include methods that operate over a user-supplied
corpus, are open or faithfully re-implementable, answer
research questions, and run on our backbone; Glia,
the AI Scientist~\cite{lu2024aiscientist,yamada2025aiscientistv2},
and AlphaEvolve~\cite{novikov2025alphaevolve} produce
algorithms or papers judged by other criteria, and
Self-RAG~\cite{asai2024selfrag} requires its own fine-tuned
generator (Appendix~A.1).

\vspace{1mm}

\textbf{Shared configuration.}
All systems use GPT-5.5 (temperature 0.2, at most 4,096 output tokens per call), the same 2,627-file corpus snapshot, task instructions, and persona, at most 8 tool steps per turn, an 8,000-token evidence budget where context is assembled, and a 3-minute limit per NoeVeraBench task; C1 search runs instead allow up to $k=30$ simulator calls and 20 minutes, without the step or 3-minute limits.
Structured-RAG indices use GPT-5.5-mini for extraction and text-embedding-3-large for embeddings.
Prompts and hyperparameters were tuned on the development split only: Hybrid RAG and NoeVera, which share the retrieval substrate, under a 12-iteration budget, and GraphRAG, HippoRAG~2, LightRAG, and PaperQA2 with their authors' recommended configurations plus 3 iterations of backbone adaptation (Appendix~A.2).
All systems were frozen before the held-out split was unsealed.
Every task starts from the same snapshot $\mathcal{B}_t$ with the Agent Memory Plane excluded from retrieval, and write-back is disabled except in \S\ref{sec:res_c2}.

\vspace{1mm}

\textbf{NoeVeraBench.}
The benchmark has 72 tasks in three tracks.
\emph{Knowledge Navigation} (N, 25 tasks) asks systems to locate, explain, synthesize, and critique evidence; it includes 8 pure literature-synthesis tasks, where literature agents and graph retrieval should be competitive, and 6 calibration tasks whose correct answer is that the evidence does not support a numerical claim or general recommendation.
\emph{Evidence Evolution} (E, 22 tasks) pairs a recommendation with a controlled evidence update that either invalidates it (11 tasks) or leaves it intact (11); the system must revise or retain its decision and justify the choice.
\emph{Decision-and-Experiment} (D, 25 tasks) is diagnostic only, because its quantitative references come from Vidur, the simulator used at evaluation time.
61\% of tasks reference recognizable serving stacks (vLLM, TensorRT-LLM, Orca, or Sarathi); they span five hardware classes, each of six core serving mechanisms appears in at least 8 tasks, and the required evidence ranges over papers, code, documentation, benchmark reports, and telemetry (Appendix~C.1).
Domain experts wrote 38 tasks, 12 were adapted from anonymized usage, and 22 are controlled evidence updates; four systems researchers reviewed and annotated every task (Appendix~C.2).
The 24-task development split (8 per track) was used for tuning; the remaining 48 tasks were held out for evaluation, comprising 33 tasks sealed before system freezing and 15 written by three outside researchers after freezing.
Two blinded LLM judges from different model families score each response against its anchors, and a third grader adjudicates large disagreements; scores are length-controlled at 1,240 words by a mixed-effects model (adjustment below 0.01; no ranking changes).
On a stratified 12.5\% sample, human graders agree at Krippendorff's $\alpha\ge0.79$ on N and E, judge--human $\alpha$ is 0.78--0.80, the judges correlate at Spearman 0.88--0.89, and a rubric audit removed six items that rewarded tool use rather than correct evidence use (Appendix~C.3).

\textbf{Statistics.}
Each system is run with 3 seeds on the held-out split; we report means with 95\% confidence intervals from a hierarchical bootstrap over tasks and runs and test paired differences with permutation tests.
The confirmatory family contains exactly two comparisons, corrected with Holm--Bonferroni: evidence records versus Mem0 on over-generalization (C2) and NoeVera without Vidur versus the best prompt-matched baseline on evidence-evolution score (C3).
C1 has no confirmatory test: with four scenarios, the smallest attainable two-sided Wilcoxon $p$-value is 0.125, so we report per-scenario consistency.
All $p$-values are raw and two-sided (Appendix~D).

\begin{table*}[t]
\centering
\caption{Configuration search (C1, diagnostic). (a) Calls-to-5\%: median [IQR] over 20 runs (4 scenarios $\times$ 5 seeds) and per-scenario medians; $r(10)\,|\,$feas.: mean regret at $k=10$ over runs that found a feasible configuration (their share: Feasible; Appendix~E.1). (b) vLLM on A100 GPUs, best configuration at $k=10$: throughput (Eq.~\ref{eq:objective}) relative to the vLLM default, averaged over feasible seeds; TTFT$_{p95}$ of the median-TTFT feasible-seed configuration ($^{\times}$: SLO violated; random/grid TTFTs recorded in the experiment logs). NoeVera+G searches with GAUSS but, like every row, is scored by Vidur against the exhaustive Vidur optimum (the regret-0 reference; 1.34$\times$ measured throughput; Appendix~B). Means use unrounded per-scenario values.}
\vspace{2mm}
\label{tab:search}
\small
\setlength{\tabcolsep}{4.8pt}
\textit{(a) Search efficiency in the simulator}\\[2pt]
\begin{tabular}{l c cccc c c}
\toprule
Method & Calls-to-5\% $\downarrow$ & Sc1 & Sc2 & Sc3 & Sc4 & $r(10)\,|\,$feas.\ $\downarrow$ & Feasible $\uparrow$ \\
\midrule
vLLM default (no search)       & ---          & --- & --- & --- & --- & 0.087 & --- \\
Random search                  & 12 [8, 18]   & 14 & 11 & 12 & 10 & 0.038 & 80\% \\
Grid search (coarse)           & 20 [15, 28]  & 22 & 18 & 19 & 20 & 0.042 & 85\% \\
Bayesian optimization (BO)     & 9 [6, 14]    & 10 & 8  & 9  & 8  & 0.028 & 90\% \\
Vidur-Search-style             & 11 [7, 16]   & 12 & 10 & 11 & 10 & 0.033 & 85\% \\
ReAct+V (no KB)                & 15 [10, 22]  & 16 & 14 & 15 & 14 & 0.041 & 80\% \\
\nb NoeVera+V                & \textbf{6 [4, 9]} & \textbf{7} & \textbf{5} & \textbf{6} & \textbf{5} & \textbf{0.019} & \textbf{95\%} \\
\nb NoeVera+G (GAUSS)        & 7 [5, 11]    & 8 & 6 & 7 & 6 & 0.022 & \textbf{95\%} \\
\bottomrule
\end{tabular}

\vspace{9pt}
\textit{(b) Validation with vLLM on A100 GPUs}\\[2pt]
\setlength{\tabcolsep}{3.5pt}
\begin{tabular}{l ccccc cccc c}
\toprule
 & \multicolumn{5}{c}{Throughput relative to vLLM default ($\times$) $\uparrow$} & \multicolumn{4}{c}{TTFT$_{p95}$ in ms [SLO $\tau$] $\downarrow$} & \\
\cmidrule(lr){2-6}\cmidrule(lr){7-10}
Method & Sc1 & Sc2 & Sc3 & Sc4 & Mean & Sc1 [300] & Sc2 [500] & Sc3 [800] & Sc4 [1200] & SLO met \\
\midrule
vLLM default                   & 1.00 & 1.00 & 1.00 & 1.00 & 1.00 & 318$^{\times}$ & 402 & 480 & 390 & 3/4 \\
Random search                  & 1.14 & 1.16 & 1.13 & 1.12 & 1.14 & --- & --- & --- & --- & 3/4 \\
Grid search (coarse)           & 1.10 & 1.14 & 1.11 & 1.12 & 1.12 & --- & --- & --- & --- & 3/4 \\
Bayesian optimization (BO)     & 1.22 & 1.24 & 1.21 & 1.20 & 1.22 & 296 & 341 & 452 & 358 & 4/4 \\
Vidur-Search-style             & 1.18 & 1.20 & 1.17 & 1.19 & 1.18 & 305$^{\times}$ & 352 & 462 & 368 & 3/4 \\
ReAct+V (no KB)                & 1.14 & 1.17 & 1.15 & 1.14 & 1.15 & 312$^{\times}$ & 368 & 472 & 378 & 3/4 \\
\nb NoeVera+V                & \textbf{1.30} & \textbf{1.34} & \textbf{1.31} & \textbf{1.29} & \textbf{1.31} & \textbf{287} & \textbf{326} & \textbf{438} & \textbf{342} & \textbf{4/4} \\
\nb NoeVera+G (GAUSS)        & 1.28 & 1.33 & 1.30 & 1.27 & 1.30 & 291 & 330 & 442 & 346 & 4/4 \\
\bottomrule
\end{tabular}
\end{table*}

\section{Results}
\label{sec:results}

\subsection{C1: Knowledge-Guided Configuration Search}
\label{sec:res_c1}

\textbf{Setup.}
Each method searches four scenarios with Llama-3-8B and Llama-3-70B on A100-80GB GPUs (Appendix~A.3) with up to $k=30$ calls and 5 seeds per scenario (20 runs per method).
Regret is $r(k)=1-f(\hat{c}_k)/f^{*}$ against the exhaustive-Vidur optimum, where $f$ is Vidur-estimated throughput and $\hat{c}_k$ the best feasible configuration within $k$ calls; \emph{calls-to-5\%} is the smallest $k$ with $r(k)\le0.05$ (runs never reaching it would be censored at 30; none were).
Each method's best configuration at $k=10$, the benchmark budget, is measured with vLLM on A100 GPUs, with TTFT$_{p95}$ at the scenario's offered load.
Runs with no feasible configuration within 10 calls (1 of 20 for each NoeVera variant, 2--4 for every baseline) are excluded from the throughput mean, and TTFT and SLO compliance come from the median-TTFT feasible-seed configuration.
Feasibility is judged on \emph{simulator} predictions (all 19 feasible runs of each NoeVera variant meet the SLO in simulation) and does not by itself certify hardware compliance.
Baselines are random search, coarse grid search, Bayesian optimization (BO), a Vidur-Search-style procedure, and the knowledge-free ReAct+V agent.

\textbf{Prior knowledge differs.}
Only the NoeVera variants can consult the offline Vidur grid, so this is a full-system comparison in which accumulated evidence (including grid estimates that may already cover these scenarios) and the search policy contribute jointly; removing the grid raises NoeVera+V's calls per benchmark query from 1.6 to 3.4 (Appendix~E.5), and a matched-prior experiment is future work.

\textbf{Results.}
NoeVera+V comes within 5\% of the optimum in a median of 6 calls (IQR 4--9), versus 9 for BO, 11 for the Vidur-Search-style procedure, 12 for random search, 15 for ReAct+V, and 20 for grid search (Table~\ref{tab:search}a).
It has the lowest median in every scenario (consistency, not significance, with four scenarios), finds a feasible configuration within 10 calls in 19 of 20 runs, and has the lowest conditional regret at $k=10$.
On hardware (Table~\ref{tab:search}b), its configurations are the fastest in every scenario (1.29--1.34$\times$ the vLLM default; mean 1.31$\times$, versus 1.34$\times$ for the exhaustive optimum) with the lowest TTFT.
The TTFT constraint binds only in Sc1 ($\tau=300$~ms), where NoeVera+V, NoeVera+G, and BO meet it and the default, random, grid, Vidur-Search-style, and ReAct+V configurations do not; elsewhere TTFTs clear the SLOs by 98--858~ms, so both NoeVera variants meet the SLO on hardware in all four scenarios.
Within each scenario, Vidur's predictions for the six Vidur-based methods' configurations rank-correlate with measured throughput at a median Spearman $\rho$ of 0.83 (range 0.77--0.94; pooled 0.87, inflated by the 8B--70B gap).
Vidur understates absolute gaps, and unevenly: it places the default 8.7\% below the optimum versus 25\% measured (plausibly operator-profiler calibration), and the baselines' measured gaps (9--16\%) are three to four times their simulated regrets at $k=10$ (2.8--4.2\%), whereas NoeVera+V's are close (2.2\% vs.\ 1.9\%).
We therefore use the simulator only to rank configurations and base all hardware claims on vLLM measurements.

\textbf{Swapping the on-demand simulator.}
NoeVera+G replaces the on-demand Vidur with GAUSS~\cite{bharadwaj2026gauss} but keeps the offline grid, and its configurations are scored by Vidur against the same optimum, so only the on-demand estimator changes.
It needs one more median call than NoeVera+V in every scenario (7 [5, 11] vs.\ 6 [4, 9]) but fewer than every baseline, all of which search with Vidur.
GAUSS rank-correlates slightly less well with measured throughput (median $\rho$ 0.80, range 0.71--0.89), yet NoeVera+G's configurations reach 1.30$\times$ the vLLM default (vs.\ 1.31$\times$) and meet all four SLOs with TTFT 4~ms higher (Appendix~B).

\subsection{C1 Diagnostics: When to Simulate and What It Adds}
\label{sec:res_inv}

\begin{table}[t]
\centering
\vspace{-2mm}
\caption{When to experiment (C1, diagnostic), on a separate 48-task invocation set (3 seeds). $Q_{\mathrm{opp}}$/$Q_{\mathrm{rel}}$: judge-scored quality on the 12 opportunity tasks / 36 no-call controls; $P_{\mathrm{tool}}$/$R_{\mathrm{tool}}$: invocation precision/recall from call logs; Unn.: unnecessary Vidur calls per control; Sim.: Vidur seconds per task. Never-call rows were measured separately for each system.}
\vspace{2mm}
\label{tab:invocation}
\footnotesize
\setlength{\tabcolsep}{3.6pt}
\begin{tabular}{l cc cc cc}
\toprule
Policy & $Q_{\mathrm{opp}}$ & $Q_{\mathrm{rel}}$ & $P_{\mathrm{tool}}$ & $R_{\mathrm{tool}}$ & Unn.\ $\downarrow$ & Sim. \\
\midrule
\multicolumn{7}{l}{\textit{NoeVera+V}} \\
\nb \hspace{0.5em}Never call     & 0.48 & 0.68 & ---  & 0.00 & 0    & 0 \\
\nb \hspace{0.5em}Selective      & 0.72 & 0.67 & 0.77 & 0.83 & 0.08 & 21 \\
\nb \hspace{0.5em}Always call    & 0.74 & 0.62 & 0.25 & 1.00 & 2.7  & 78 \\
\multicolumn{7}{l}{\textit{ReAct+V}} \\
\hspace{0.5em}Never call        & 0.48 & 0.68 & ---  & 0.00 & 0    & 0 \\
\hspace{0.5em}Selective          & 0.63 & 0.68 & 0.63 & 0.72 & 0.14 & 26 \\
\hspace{0.5em}Always call        & 0.66 & 0.62 & 0.25 & 1.00 & 2.9  & 82 \\
\bottomrule
\vspace{-5mm}
\end{tabular}
\end{table}

A reliable agent should also know when experimentation is unnecessary.
On a separate 48-task invocation set disjoint from NoeVeraBench (12 experiment-opportunity tasks and 36 no-call controls whose answers need no simulation), we compare never-call, always-call, and selective policies (Table~\ref{tab:invocation}).
Per seed, NoeVera+V's selective policy called Vidur on 13 tasks on average, 10 of which needed a call, taking 10 of the 12 opportunities, with 0.08 unnecessary calls per control (3 controls received one each); it retains 97\% of always-call quality on opportunity tasks at 27\% of the simulator time.
Always calling degrades both systems' no-call-control quality (0.68$\to$0.62), whereas selective invocation stays near the no-call baseline (0.67 for NoeVera+V, 0.68 for ReAct+V): unnecessary calls hurt, and selectivity prevents them.
NoeVera's controller also decides better than ReAct+V's: higher recall (0.83 vs.\ 0.72), precision (0.77 vs.\ 0.63), and $Q_{\mathrm{opp}}$ (0.72 vs.\ 0.63), with fewer unnecessary calls (0.08 vs.\ 0.14).

\textbf{What the simulator adds.}
Without a simulator, no system exceeds 0.48 on the diagnostic D track (Table~\ref{tab:e2e}; per-intent scores in Appendix~E.2), and method comparison, a comparative rather than quantitative intent, is the only one on which retrieval-based systems exceed 0.6.
Adding the same Vidur interface lifts every system, but the lift tracks agentic orchestration more than retrieval: $+0.15$ for LLM-only, $+0.13$ for Hybrid RAG, $+0.19$ for ReAct, and $+0.25$ for NoeVera, whose advantage over ReAct+V is largest on experiment design ($+0.12$) and optimization ($+0.08$).
Knowledge combined with NoeVera's orchestration thus makes the same simulator more useful, consistent with C1.

\subsection{C2: Safe Reuse of Persistent Experimental Feedback}
\label{sec:res_c2}

\textbf{Setup.}
From a fixed snapshot $\mathcal{B}_t$, each of 8 seed questions triggers a Vidur investigation whose finding is written back (\S\ref{sec:loop}).
With fresh Vidur calls disabled, we pose four kinds of follow-up; across 3 seeds, we score 36 questions to which the finding applies ($F_{\mathrm{app}}$), 36 superficially similar \emph{traps} whose GPU, model size, or workload invalidates it ($F_{\mathrm{trap}}$), 24 on which it conflicts with an existing source obtained under different conditions ($F_{\mathrm{conf}}$), and 24 unrelated controls ($F_{\mathrm{ctrl}}$).
To isolate storage format, reasoning is fixed to NoeVera's orchestration while the finding is stored as a plain chunk in the Hybrid RAG index, a Mem0~\cite{chhikara2025mem0} memory, or an evidence record; the ``ReAct+chunk'' column of Table~\ref{tab:reuse} is an end-to-end reference in which ReAct reads a plain-chunk finding through its own loop (its no-finding score is 0.68, so $+0.02$ is the gain from the finding).
We report the reuse gain $\Delta_{\mathrm{reuse}}$ (the mean change in grounded decision quality on $F_{\mathrm{app}}$ from $\mathcal{B}_t$ to $\mathcal{B}_{t+1}$) and correct-reuse, over-generalization, conflict-surfacing, and drift rates.
A response over-generalizes when it recommends the stored configuration to a trap for which the pre-recorded answer says the finding does not apply; this rule needs no LLM judge, so the metric is comparable across storage formats even though only evidence records store applicability conditions explicitly.
Tests and intervals operate at the level of the 8 parent investigations.

\begin{table}[t]
\centering
\caption{Reusing written-back findings (C2), fresh Vidur calls disabled; reasoning is fixed to NoeVera's orchestration except in the ReAct column. Reused, Over-gen., Conflict: rates on $F_{\mathrm{app}}$, $F_{\mathrm{trap}}$, $F_{\mathrm{conf}}$ ($n=36$, 36, 24; 3 seeds); Drift: mean $|\Delta Q|$ on $F_{\mathrm{ctrl}}$.}
\vspace{2mm}
\label{tab:reuse}
\footnotesize
\setlength{\tabcolsep}{3.9pt}
\begin{tabular}{l c ccc c}
\toprule
 & \hd{No}{finding} & \hd{Plain}{chunk} & \hd{Mem0}{memory} & \hd{Evidence}{record} & \hd{ReAct +}{chunk} \\
\midrule
$Q(F_{\mathrm{app}})$ $\uparrow$        & 0.68 & 0.70    & 0.71    & \cellcolor{hlrow}\textbf{0.77}    & 0.70 \\
$\Delta_{\mathrm{reuse}}$ $\uparrow$    & ---  & $+0.02$ & $+0.03$ & \cellcolor{hlrow}$\mathbf{+0.09}$ & $+0.02$ \\
Reused $\uparrow$                        & ---  & 58\%    & 61\%    & \cellcolor{hlrow}\textbf{83\%}    & 56\% \\
Over-gen.\ $\downarrow$                  & ---  & 33\%    & 28\%    & \cellcolor{hlrow}\textbf{8\%}     & 39\% \\
Conflict $\uparrow$                      & ---  & 42\%    & 50\%    & \cellcolor{hlrow}\textbf{83\%}    & 38\% \\
Drift $\downarrow$                       & ---  & 0.048   & 0.039   & \cellcolor{hlrow}\textbf{0.012}   & 0.052 \\
\bottomrule
\vspace{-6mm}
\end{tabular}
\end{table}

\vspace{2mm}

\textbf{Results.}
Evidence records over-generalize on 8\% of traps, versus 28\% for Mem0 and 33\% for plain chunks (Table~\ref{tab:reuse}).
In the C2 confirmatory comparison, a parent-level sign-flip test over the 8 parents gives raw $p=0.016$, which survives Holm correction (threshold 0.05).
Evidence records also surface 83\% of conflicts (Mem0: 50\%), reuse findings correctly on 83\% of applicable follow-ups, barely perturb unrelated controls (drift 0.012), and raise applicable-follow-up quality by $+0.09$ versus $+0.03$ for Mem0 (difference $+0.06$, 95\% CI $[0.03,0.10]$; secondary outcome).

\begin{figure*}[t]
\centering
\includegraphics[width=0.9\linewidth]{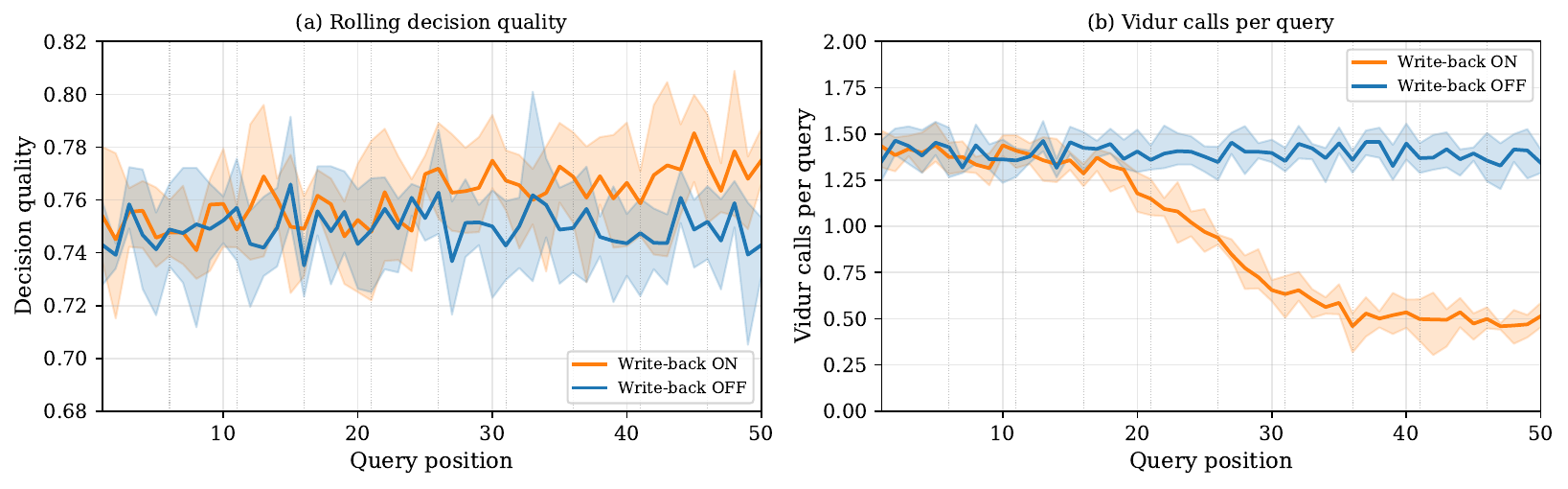}
\caption{Accumulation over a 50-query stream with write-back on and off: (a) decision quality and (b) Vidur calls per query (means over 5 random orderings; shaded: range; dotted lines: first query of each 5-query cluster).}
\label{fig:stream}
\end{figure*}

\textbf{Accumulation over a query stream.}
Over 50 related queries (10 clusters, 5 random orders) with fresh Vidur calls allowed, write-back reduces Vidur calls per query from 1.4 over the first 10 queries to 0.5 over the last 10 ($-64\%$), with savings substantial only after about 20 queries, once accumulated evidence covers later questions; quality over the stream is slightly higher with write-back ($+0.012$ mean ON$-$OFF gain, 95\% CI $[0.003,0.021]$). With write-back disabled, the streams stay at 1.4 calls per query in both windows, so the reduction comes from accumulated evidence rather than ordering or question difficulty (Figure~\ref{fig:stream}).

\subsection{C3: Research Reasoning on NoeVeraBench}
\label{sec:res_c3}

\begin{table}[t]
\centering
\caption{Normalized rubric scores on the held-out split (3 seeds; 95\% CI half-widths 0.02--0.03 for every entry, Appendix~E.3). Overall is the mean of N and E; $^{\dagger}$D is diagnostic, excluded from Overall, and the task-count-weighted mean over five intents (Appendix~E.2); E scores are descriptive here (confirmatory test in Table~\ref{tab:evo}); $^{\ast}$run on N only. All rows were measured directly; OpenHands was run only with Vidur. 
\vspace{2mm}
\textbf{Bold}: best in column.}
\label{tab:e2e}
\footnotesize
\setlength{\tabcolsep}{3.6pt}
\begin{tabular}{l cccc c}
\toprule
System & N & D$^{\dagger}$ & E & Overall & \$/query \\
\midrule
\multicolumn{6}{l}{\textit{Without Vidur}} \\
LLM-only                  & 0.47 & 0.33 & 0.31 & 0.39 & 0.02 \\
Dense RAG                 & 0.73 & 0.45 & 0.36 & 0.55 & 0.08 \\
Hybrid RAG                & 0.78 & 0.46 & 0.41 & 0.60 & 0.10 \\
GraphRAG                  & 0.83 & 0.47 & 0.45 & 0.64 & 0.09 \\
HippoRAG 2                & 0.81 & 0.46 & 0.44 & 0.63 & 0.10 \\
LightRAG                  & 0.80 & 0.46 & 0.43 & 0.62 & 0.09 \\
PaperQA2$^{\ast}$         & \textbf{0.87} & --- & --- & --- & 0.22 \\
ReAct                     & 0.82 & 0.47 & 0.46 & 0.64 & 0.18 \\
\nb NoeVera             & 0.85 & 0.48 & 0.71 & 0.78 & 0.21 \\
\midrule
\multicolumn{6}{l}{\textit{With Vidur}} \\
LLM-only+V                & 0.46 & 0.48 & 0.30 & 0.38 & 0.06 \\
Hybrid RAG+V              & 0.77 & 0.59 & 0.40 & 0.59 & 0.26 \\
ReAct+V                   & 0.80 & 0.66 & 0.44 & 0.62 & 0.31 \\
OpenHands+V               & 0.81 & 0.65 & 0.45 & 0.63 & 0.29 \\
\nb NoeVera+V           & 0.86 & \textbf{0.73} & \textbf{0.73} & \textbf{0.80} & 0.35 \\
\bottomrule
\vspace{-5mm}
\end{tabular}
\end{table}

Every corpus-grounded system improves navigation over LLM-only (Table~\ref{tab:e2e}; 0.73--0.87 versus 0.47), and PaperQA2 is strongest on N (0.87, versus 0.86 for NoeVera+V), as expected for literature synthesis: NoeVera's claims concern decisions, revision, and reuse rather than retrieval.
The separation appears on evidence evolution, where NoeVera reaches 0.71 versus 0.41 for Hybrid RAG, 0.45 for the best structured RAG, and 0.46 for ReAct.
Overall, NoeVera+V scores 0.80 versus 0.62 for the equally equipped ReAct+V and 0.59 for the single-call retrieval counterpart Hybrid RAG+V; \S\ref{sec:res_evo} shows that most of the corresponding no-Vidur E-track gap closes under prompt matching, leaving a significant residual.
The +V baselines score slightly below their non-Vidur counterparts (overall, ReAct 0.64\,$\to$\,0.62, Hybrid RAG 0.60\,$\to$\,0.59, LLM-only 0.39\,$\to$\,0.38); each drop is within the 95\% CIs, but all three point the same way, and we attribute them to Vidur calls on non-quantitative tasks displacing correct retrieved evidence.
NoeVera+V, whose controller gates calls on the query profile, keeps N and E at or slightly above their non-Vidur levels; always calling would depress its no-call-control quality from 0.68 to 0.62 (Table~\ref{tab:invocation}).
On the D track, NoeVera+V reaches 0.73 versus 0.66 for ReAct+V ($+0.07$, 95\% CI $[0.03,0.10]$; exploratory).

\subsection{C3: Revising Decisions When Evidence Changes}
\label{sec:res_evo}

To test whether the E-track advantage survives prompt engineering, two prompt-matched baselines receive NoeVera's final-synthesis instructions verbatim, including its revision template: Hybrid RAG-PM in a single call and ReAct-PM as the agent's final-answer instructions.
An oracle condition gives every system the reference turn-1 recommendation and scores only turn 2, separating revision from first-answer quality.
Table~\ref{tab:evo} covers the 14 held-out E tasks (7 revision-expected, 7 retention-expected; all 22 tasks in Appendix~E.4, with the same pattern).

\begin{table}[t]
\centering
\caption{Decision revision on the 14 held-out E tasks (3 seeds); -PM rows are prompt-matched variants of the row above. T1/T2: scores before/after the evidence update (T2 equals E in Table~\ref{tab:e2e}). Revised, Retained, and Reversed (unwarranted; lower is better): rates over 21 decisions each. Oracle: T2 given the reference turn-1 answer.}
\vspace{2mm}
\label{tab:evo}
\footnotesize
\setlength{\tabcolsep}{2.4pt}
\begin{tabular}{l cc ccc c}
\toprule
System & T1 & T2 & Revised & Retained & Reversed & Oracle \\
\midrule
LLM-only               & 0.38 & 0.31 & 19\% & 43\% & 38\% & 0.35 \\
GraphRAG               & 0.44 & 0.45 & 38\% & 57\% & 24\% & 0.49 \\
Hybrid RAG             & 0.42 & 0.41 & 33\% & 52\% & 29\% & 0.44 \\
\hspace{0.5em}-PM      & 0.46 & 0.61 & 57\% & 71\% & 14\% & 0.60 \\
ReAct                  & 0.46 & 0.46 & 43\% & 62\% & 19\% & 0.50 \\
\hspace{0.5em}-PM      & 0.48 & 0.63 & 62\% & 76\% & 10\% & 0.62 \\
\nb NoeVera          & \textbf{0.68} & 0.71 & 76\% & 86\% & \textbf{5\%} & \textbf{0.71} \\
\nb NoeVera+V        & \textbf{0.68} & \textbf{0.73} & \textbf{81\%} & \textbf{90\%} & \textbf{5\%} & \textbf{0.71} \\
\bottomrule
\end{tabular}
\end{table}

Prompt matching closes about two-thirds of the gap: Hybrid RAG rises from 0.41 to 0.61 and ReAct from 0.46 to 0.63, against 0.71 for NoeVera.
With oracle turn-1 answers the scores are 0.60, 0.62, and 0.71, so the remaining advantage reflects revision itself rather than a better first answer.
NoeVera's residual $+0.08$ over the best prompt-matched baseline (95\% CI $[0.03,0.13]$, raw $p=0.004$) is the C3 confirmatory test and survives Holm correction (threshold 0.025).
NoeVera also reverses retention-expected decisions in 5\% of cases, versus 10\% for ReAct-PM and 29\% for Hybrid RAG.
C3 thus holds in a narrowed form: a specified revision protocol recovers most of the gap, and NoeVera's knowledge and orchestration recover the rest, consistent with the 0.10--0.12 cost of replacing the orchestration over the same knowledge base (Table~\ref{tab:attrib}c); Appendix~E.4 explains two scoring details of Table~\ref{tab:evo}.

\subsection{What Each Component Contributes}
\label{sec:res_attr}

\begin{table}[t]
\centering
\caption{Attribution with Vidur and a fixed evidence budget: (a) the four combinations of authority and synthesis signals; (b, c) one factor varied at a time. Recall: annotated required evidence present in the final pack (N tasks); Pack: mean tokens. D weighted as in Table~\ref{tab:e2e}.}
\vspace{3mm}
\label{tab:attrib}
\footnotesize
\setlength{\tabcolsep}{3.4pt}
\resizebox{\linewidth}{!}{
\begin{tabular}{l cccc cc}
\toprule
Configuration & N & D & E & Overall & Recall & Pack \\
\midrule
\nb NoeVera+V            & \textbf{0.86} & \textbf{0.73} & \textbf{0.73} & \textbf{0.80} & \textbf{91\%} & \textbf{2.5k} \\
\multicolumn{7}{l}{\textit{(a) Representation}} \\
\hspace{0.5em}Evidence only (no authority) & 0.83 & 0.63 & 0.58 & 0.71 & 70\% & 3.0k \\
\hspace{0.5em}Evidence + authority          & 0.84 & 0.66 & 0.62 & 0.73 & 76\% & 2.8k \\
\hspace{0.5em}Evidence + synthesis          & 0.85 & 0.68 & 0.67 & 0.76 & 84\% & 2.7k \\
\multicolumn{7}{l}{\textit{(b) Knowledge backend supplying the context}} \\
\hspace{0.5em}GraphRAG              & 0.84 & 0.66 & 0.62 & 0.73 & 78\% & 3.0k \\
\hspace{0.5em}HippoRAG 2            & 0.83 & 0.65 & 0.60 & 0.72 & 76\% & 2.9k \\
\hspace{0.5em}LightRAG              & 0.82 & 0.64 & 0.59 & 0.71 & 75\% & 2.9k \\
\multicolumn{7}{l}{\textit{(c) Orchestration over the full KB, with Vidur}} \\
\hspace{0.5em}Single call           & 0.80 & 0.60 & 0.55 & 0.68 & 88\% & 2.5k \\
\hspace{0.5em}Tool-using agent      & 0.82 & 0.63 & 0.58 & 0.70 & 89\% & 2.6k \\
\bottomrule
\vspace{-10mm}
\end{tabular}
}
\end{table}

Over an Evidence-Plane-only knowledge base (Table~\ref{tab:attrib}a; 0.71, 70\% recall), authority adds $+0.02$ alone and synthesis $+0.05$ alone, but together $+0.09$ (to 0.80 at 91\% recall), more than the sum of their parts, while packs shrink from 3,000 to 2,500 tokens.
Feeding NoeVera's orchestration with GraphRAG, HippoRAG~2, or LightRAG context instead costs 0.07--0.09 overall and 13--16 points of recall.
Replacing the orchestration costs 0.12 (single call) or 0.10 (generic tool-using agent) despite nearly identical recall (88--89\%), so orchestration matters most where evidence must be weighed and revised.
In leave-one-out ablations (Appendix~E.5), removing any single component (a retrieval-lane group, authority ranking, evidence packing, coverage checks, persona profiling, or multi-role agents) costs at most 0.03, so the gains come from components acting together; removing authority ranking (0.77) and adding synthesis to the Evidence-only base (0.76) reach the same system by opposite constructions and agree within the CI.
Removing the offline grid barely changes scores ($-0.01$) but raises Vidur calls per query from 1.6 to 3.4: proactive enrichment mainly buys efficiency, the effect C1 measures.
Because the study mixes factor- and combination-level rows, these differences are descriptive, not a formal interaction estimate.

\subsection{Cost and Robustness}
\label{sec:res_cost}

NoeVera+V uses 78k tokens and costs \$0.35 per query (3.5$\times$ Hybrid RAG's cost, 1.1$\times$ ReAct+V's), with a median latency of 71~s (p95 118~s), 31\% of it in Vidur running in parallel with the investigation.
A compute-matched Hybrid RAG with the same 78k tokens (best-of-3 self-consistency and a larger retrieval budget) reaches only 0.63, so the gain is not explained by extra LLM computation, although matching tokens does not equalize simulator time or offline compute.
Building the Second Brain costs \$294 (mostly LLM distillation and synthesis) and 23.3~hours (mostly the offline Vidur grid), or \$0.64 per query amortized over $10^3$ queries and \$0.38 over $10^4$, versus \$0.11--0.28 and \$0.10--0.11 for the baselines' indices.
The $+0.18$ Overall gap over ReAct+V (95\% CI $[0.13,0.23]$) holds for outside-authored tasks ($+0.14$) and an alternate grader ($+0.16$), and narrows with weaker backbones while its 95\% CI still excludes zero ($+0.11$ with GPT-4o, $+0.09$ with Llama-3.3-70B; Appendix~E.6).

\section{Related Work}
\label{sec:related}

\textbf{Retrieval, knowledge graphs, and agent memory.}
Self-RAG~\cite{asai2024selfrag} and Corrective RAG~\cite{yan2024crag} make retrieval adaptive and self-correcting; RAPTOR~\cite{sarthi2024raptor} retrieves over hierarchical summaries; GraphRAG~\cite{edge2024graphrag}, LightRAG~\cite{guo2025lightrag}, and HippoRAG~\cite{gutierrez2024hipporag,gutierrez2025hipporag2} add entity graphs, dual-level retrieval, and associative recall; and AgenticRAG~\cite{suresh2026agenticrag} gives a reasoning model iterative search and evidence-analysis tools.
Generative Agents~\cite{park2023generative}, MemGPT~\cite{packer2023memgpt}, Mem0~\cite{chhikara2025mem0}, and A-MEM~\cite{xu2025amem} let agents accumulate memories beyond a context window.
NoeVera builds on these mechanisms but targets decision-ready technical evidence: it separates sources from synthesis and usage records, ranks by authority, applicability, and lifecycle, and stores experimental findings as scoped evidence rather than conversational memory, the distinction behind its over-generalization gap against Mem0 (\S\ref{sec:res_c2}).
Unlike retrieval systems focused primarily on finding relevant information, NoeVera also tracks the conditions under which experimental evidence remains applicable to subsequent decisions.

\vspace{2mm}

\textbf{Agents for research and systems optimization.}
PaperQA2~\cite{skarlinski2024paperqa2} and OpenScholar~\cite{asai2026openscholar} synthesize scientific literature with citation-grounded generation; the AI Scientist~\cite{lu2024aiscientist,yamada2025aiscientistv2} automates the path from idea generation to manuscript; AstaBench~\cite{bragg2026astabench} benchmarks research agents; AlphaEvolve~\cite{novikov2025alphaevolve} evolves programs against automated evaluators; and KernelBench~\cite{ouyang2025kernelbench} tests GPU-kernel generation.
Closest to our work, Glia~\cite{hamadanian2025glia} coordinates Researcher and Supervisor agents that form hypotheses, run experiments, and refine system designs, showing that reasoning-guided experimentation can outperform less structured search.
Glia asks whether agents can \emph{discover} better mechanisms; NoeVera asks how agents can \emph{accumulate, navigate, and reuse} evidence across questions.
The two are complementary, as a Second Brain could supply reusable knowledge to an autonomous design agent, and we make no claim to discover better algorithms.
NoeVera further closes the knowledge--experiment loop by using retrieved evidence to guide simulator calls and writing back their results for subsequent reasoning.

\textbf{LLM inference simulation.}
Vidur~\cite{agrawal2024vidur} combines operator profiling with predictive modeling and, through Vidur-Search, automates deployment-configuration search; LLMServingSim~\cite{cho2024llmservingsim,cho2026llmservingsim2} co-simulates hardware and software for homogeneous, heterogeneous, and disaggregated serving; and GAUSS~\cite{bharadwaj2026gauss} trades Vidur's operator profiling for an analytical + Monte-Carlo approach with lower per-call cost.
NoeVera introduces no new simulation method and treats the simulator as a swappable component.
Its contribution lies in selecting which configurations warrant simulation using accumulated technical evidence, rather than improving the simulator's predictive model.

\section{Discussion and Limitations}
\label{sec:discussion}

\textbf{Lessons.}
Our evaluation yields three broader lessons for
knowledge-guided research systems.
First, knowledge organization is a first-class design
choice: separating evidence from synthesis lets agents
navigate compact representations without losing the
conditions behind a conclusion, while scoped evidence
records preserve applicability conditions to prevent
written-back findings from being over-generalized
(\S\ref{sec:res_c2}).
Second, relevant evidence alone is insufficient for
decision-making: orchestration helps most when evidence
must be weighed and revised, but structure can't
substitute for quantitative measurement, as no system
exceeded 0.48 on track D without a simulator.
Knowing \emph{when} to experiment is therefore distinct
from knowing what is already known.
Unnecessary experiments also carries a cost:
the +V baselines perform slightly worse on N and E
(\S\ref{sec:res_c3}), and always invoking the simulator
reduces no-call-control quality from 0.68 to 0.62
(Table~\ref{tab:invocation}).
Third, experiments should extend knowledge rather than
end as isolated tool calls: persisting findings with
their applicability conditions reduces subsequent
simulation without degrading answer quality.

\textbf{Where NoeVera does not help, and limitations.}
A dedicated literature-synthesis agent matches NoeVera on Knowledge Navigation (PaperQA2: 0.87 versus 0.86); without a simulator, NoeVera is no better than plain retrieval or a generic agent on the D track (0.48); and it costs 3.5$\times$ as much per query as single-call RAG and needs an offline build, a trade-off most favorable when related questions recur (\S\ref{sec:res_c2}).
In configuration search, only the NoeVera variants have the offline grid, so their gains combine prior evidence and search policy; separating them requires a matched-prior variant.
Four scenarios support consistency but not significance for C1, and the C2 test rests on 8 parent investigations.
The backend swap tests sensitivity to the on-demand estimator, not independence from Vidur, since NoeVera+G keeps the Vidur-built grid and Vidur's optimum as its target.
Simulator estimates inherit their modeling assumptions (Vidur's understate absolute gaps, unevenly), and provenance is not correctness: a traced source can still be wrong, and persistence alone does not guarantee that the right evidence is retrieved later.
Also, NoeVeraBench's 72 tasks are a bounded sample of LLM-systems research, and performance remains sensitive to the backbone model.
Finally, our implementation targets LLM serving, where structured evidence and performance models are readily available; other domains may need different representations, authority criteria, and tools, with benefits weighed against offline and online costs.

\section{Conclusion}
\label{sec:conclusion}

NoeVera treats knowledge in LLM-serving research as a persistent, provenance-preserving system rather than a collection of documents retrieved anew for each question. By separating evidence, synthesis, and agent memory, navigating for decisions rather than similarity alone, and writing scoped experimental findings back into the knowledge base, NoeVera reaches strong configurations with fewer simulator calls (with Vidur or GAUSS), reuses past findings without over-generalizing them, and revises decisions when evidence changes. 
We see this evidence-to-decision loop as a foundation for research agents that understand what is known, what remains uncertain, and what is worth investigating next.

\clearpage

\appendix

\onecolumn  %

\section{Systems and Experimental Configuration}
\label{app:systems}

\subsection{Inclusion Criteria}

We benchmark a method if it (i) operates over a user-supplied corpus, (ii) is open source or faithfully re-implementable, (iii) produces the same kind of output as NoeVera (answers to research questions and decisions), and (iv) can run on our backbone.
Glia~\cite{hamadanian2025glia}, the AI Scientist~\cite{lu2024aiscientist,yamada2025aiscientistv2}, and AlphaEvolve~\cite{novikov2025alphaevolve} produce algorithms or papers judged by different criteria and so fail (iii); the configuration-search experiment (\S4.1) instead compares NoeVera with optimization-style search.
Vidur~\cite{agrawal2024vidur}, GAUSS~\cite{bharadwaj2026gauss}, and LLMServingSim~\cite{cho2024llmservingsim} are quantitative tools rather than competing systems, and a Vidur-Search-style procedure appears as a search baseline.
Self-RAG~\cite{asai2024selfrag} requires its own fine-tuned generator and fails (iv).
Commercial deep-research products search the open web rather than our corpus, cannot call a simulator, and change over time, so they fail (i) and (ii); we do not compare against them.

\subsection{Per-System Configuration}

\vspace{-3mm}

\begin{table}[H]
\centering
\caption{Per-system configuration and tuning. All systems use GPT-5.5 (temperature 0.2, at most 4,096 output tokens per call), an 8,000-token evidence budget where context is assembled, and at most 8 tool steps per turn. All tuning used the development split only: Hybrid RAG and NoeVera, which share the hybrid retrieval substrate, under a 12-iteration budget; GraphRAG, HippoRAG 2, LightRAG, and PaperQA2 keep their authors' recommended configurations with 3 iterations of backbone adaptation. The last column gives the iterations used. NoeVera+G shares the prompts and configuration of NoeVera+V, so its tuning iterations are inherited.}
\vspace{2mm}
\label{tab:config}
\small
\resizebox{0.9\textwidth}{!}{
\begin{tabular}{l p{0.64\linewidth} c}
\toprule
System & Configuration & Tuning iterations \\
\midrule
LLM-only      & Single call, no retrieval & 0 \\
Dense RAG     & OpenAI text-embedding-3-large, $k=10$, cosine similarity & 4 \\
Hybrid RAG    & NoeVera's hybrid retrieval lanes and authority- and lifecycle-aware ranking over the raw files together with the Evidence Plane's source representations and authority metadata, without the Synthesis and Agent Memory planes; single call & 12 \\
GraphRAG      & Authors' defaults; GPT-5.5-mini extraction; GPT-5.5 synthesis & 3 \\
HippoRAG 2    & Authors' defaults & 3 \\
LightRAG      & Authors' defaults & 3 \\
PaperQA2      & Authors' defaults (N track only) & 3 \\
ReAct         & Hybrid search over raw files, document reader, ReAct loop, 8 steps & 8 \\
ReAct+V       & ReAct plus Vidur through the same function-call interface as NoeVera+V & 10 \\
OpenHands+V   & OpenHands with the raw file tree, \texttt{grep}, and the Vidur CLI, 8 steps & 6 \\
LLM-only+V    & Single call, Vidur through function calling & 3 \\
Hybrid RAG+V  & Hybrid RAG plus Vidur through function calling & 6 \\
NoeVera     & Evidence and Synthesis planes (Agent Memory not retrieved), hybrid retrieval lanes, authority ranking, evidence packing, controller with scout, investigator, and synthesizer roles & 12 \\
NoeVera+V   & NoeVera plus Vidur (interface and documentation identical to ReAct+V) & 12 \\
NoeVera+G   & NoeVera plus GAUSS as the on-demand backend; offline grid unchanged; prompts and configuration inherited from NoeVera+V & 12 (inherited) \\
\bottomrule
\end{tabular}
}
\end{table}

\subsection{Configuration-Search Protocol and Scenarios}
\label{app:search}

The protocol, metrics, and results are in \S4.1. Two conventions are not stated there: a run with no feasible configuration within $k$ calls is assigned $r(k)=1$, which is why Table 1(a) reports $r(10)$ conditional on feasibility (counts in Table~\ref{tab:feasibility}); and the pooled Spearman $\rho$ of 0.87 is computed over the 24 (method, scenario) means over the six Vidur-based methods, excluding the default and the exhaustive optimum.

\textbf{Vidur setup.}
Vidur reports less than 9\% latency error~\cite{agrawal2024vidur}.
Its supported model and device matrix lists Meta-Llama-3-8B and Meta-Llama-3-70B on A100-80GB DGX and 4$\times$A100-80GB pairwise-NVLink nodes, but not on H100 DGX, which is why every scenario uses A100 GPUs.
Prefix-caching configurations were simulated with the prefix-caching implementation on the \texttt{canary} branch of \texttt{microsoft/vidur} (commit recorded in the harness).
Sc4's 4,224-token requests exceed Vidur's default 4k context, so Sc4 runs pass the three extended-prediction parameters documented in the repository; Sc1 (640 tokens), Sc2 (2,304), and Sc3 (1,536) fit within 4k:
\begin{flushleft}\scriptsize\ttfamily
--random\_forrest\_execution\_time\_predictor\_config\_prediction\_max\_prefill\_chunk\_size 16384\\
--random\_forrest\_execution\_time\_predictor\_config\_prediction\_max\_batch\_size 512\\
--random\_forrest\_execution\_time\_predictor\_config\_prediction\_max\_tokens\_per\_request 16384
\end{flushleft}

\textbf{GAUSS setup.}
GAUSS is an analytical + Monte-Carlo simulator~\cite{bharadwaj2026gauss}. We used its default supported configurations for Meta-Llama-3-8B and Meta-Llama-3-70B on A100-80GB hardware. The 20-minute per-run cap did not bind for either backend, so the calls-to-5\% comparison isolates the effect of the on-demand estimator.

\begin{table}[H]
\centering
\caption{Configuration-search scenarios, all on A100-80GB GPUs. Workloads give prompt/output tokens and the arrival process. The worked example in \S2.4 uses Sc1 with the TTFT limit relaxed from 300 to 400~ms.}
\vspace{3mm}
\label{tab:scenarios}
\small
\resizebox{0.9\textwidth}{!}{
\begin{tabular}{l l l l c c}
\toprule
Scenario & Model & Hardware & Workload & TTFT$_{p95}$ limit $\tau$ & Memory $M_h$ \\
\midrule
Sc1 & Llama-3-70B BF16, TP=4 & 4$\times$A100-80GB pairwise & 512/128, Poisson 8 rps  & 300~ms  & 76~GB \\
Sc2 & Llama-3-8B BF16, TP=1  & 1$\times$A100-80GB          & 2048/256, Poisson 16 rps & 500~ms  & 76~GB \\
Sc3 & Llama-3-70B BF16, TP=4 & 4$\times$A100-80GB pairwise & 1024/512, burst arrivals & 800~ms  & 76~GB \\
Sc4 & Llama-3-8B BF16, TP=1  & 1$\times$A100-80GB          & 4096/128, Poisson 4 rps  & 1200~ms & 76~GB \\
\bottomrule
\end{tabular}
}
\end{table}

\section{GAUSS Simulator-Backend Study}
\label{app:gauss}

The main paper reports C1 with Vidur as the on-demand backend.
To test the workflow's sensitivity to the choice of on-demand simulator, we repeated the same protocol with GAUSS~\cite{bharadwaj2026gauss}, an analytical + Monte-Carlo simulator whose estimation approach differs from Vidur's operator profiling.
NoeVera+G uses the same provenance-tagged, Vidur-generated offline records as prior evidence, and the same seeds, prompts, and persona; only the on-demand backend changes.
GAUSS shares the same arrival-rate-sweep capacity definition and models prefix caching; Sc4's extended context is handled as for Vidur.
Vidur scoring of GAUSS-selected configurations is strictly evaluation-only and never fed back into NoeVera+G's search. Feasibility and regret in Table~\ref{tab:gauss_per_scenario} are scored against Vidur's predictions since the exhaustive optimum is Vidur-defined, and the calls-to-5\% metric counts only GAUSS's online evaluations.

\begin{table}[H]
\centering
\caption{GAUSS vs. Vidur backend, per-scenario. Left: median calls-to-5\% per scenario. Right: mean $r(10)$ among feasible runs per scenario. Both backends share the same offline prior and the same evaluation protocol; only the on-demand estimator differs. Feasible: runs that found a feasible configuration within $k=10$ calls.}
\vspace{3mm}
\label{tab:gauss_per_scenario}
\small
\resizebox{0.65\textwidth}{!}{
\begin{tabular}{l cc cc c}
\toprule
 & \multicolumn{2}{c}{Median calls-to-5\%} & \multicolumn{2}{c}{$r(10)$ among feasible} & Feasible \\
\cmidrule(lr){2-3}\cmidrule(lr){4-5}
Scenario & Vidur (+V) & GAUSS (+G) & Vidur (+V) & GAUSS (+G) & (+G) \\
\midrule
Sc1 & 7  & 8  & 0.021 & 0.025 & 5/5 \\
Sc2 & 5  & 6  & 0.016 & 0.019 & 5/5 \\
Sc3 & 6  & 7  & 0.020 & 0.023 & 5/5 \\
Sc4 & 5  & 6  & 0.019 & 0.022 & 4/5 \\
Aggregate & 6 [4, 9] & 7 [5, 11] & 0.019 & 0.022 & 19/20 (95\%) \\
\bottomrule
\end{tabular}
}
\end{table}

GAUSS adds one median call per scenario (7/5/6/5 $\to$ 8/6/7/6) and slightly raises the conditional regret at $k=10$ (0.019 $\to$ 0.022). Under the common Vidur-based evaluation, the swap makes the search slightly more expensive, although the experiment does not isolate why.
The efficiency ordering versus baselines is unchanged: on aggregate, NoeVera+G's 7 [5, 11] still needs fewer calls than every baseline, all of which search with Vidur: Bayesian optimization (9 [6, 14]), the Vidur-Search-style procedure (11 [7, 16]), random search (12 [8, 18]), ReAct+V (15 [10, 22]), and grid search (20 [15, 28]).

\begin{table}[H]
\centering
\caption{GAUSS vs. Vidur on hardware (vLLM on A100-80GB). Throughput is the mean over feasible seeds at $k=10$, relative to the vLLM default; TTFT$_{p95}$ and SLO status come from the median-TTFT feasible-seed configuration. GAUSS runs are evaluated against the same vLLM-default baseline so that hardware claims are comparable; the exhaustive Vidur optimum is not used here.}
\vspace{3mm}
\label{tab:gauss_hw}
\small
\resizebox{0.65\textwidth}{!}{
\begin{tabular}{l cc cc c}
\toprule
 & \multicolumn{2}{c}{Throughput ($\times$)} & \multicolumn{2}{c}{TTFT$_{p95}$ (ms)} & \\
\cmidrule(lr){2-3}\cmidrule(lr){4-5}
Scenario & Vidur (+V) & GAUSS (+G) & Vidur (+V) & GAUSS (+G) & SLO (+G) \\
\midrule
Sc1 & 1.30 & 1.28 & 287 & 291 & met \\
Sc2 & 1.34 & 1.33 & 326 & 330 & met \\
Sc3 & 1.31 & 1.30 & 438 & 442 & met \\
Sc4 & 1.29 & 1.27 & 342 & 346 & met \\
Mean & 1.31 & 1.30 & --- & --- & 4/4 \\
\bottomrule
\end{tabular}
}
\end{table}

On hardware, the two backends land on configurations with comparable performance: mean throughput 1.30$\times$ (GAUSS) vs. 1.31$\times$ (Vidur) against the vLLM default, with TTFT$_{p95}$ 4~ms higher and every scenario meeting its SLO.
The within-scenario Spearman rank correlation between simulator-predicted and measured throughput is median 0.80 for GAUSS (range 0.71--0.89, computed over its predictions for the same six per-method configurations) versus 0.83 for Vidur (range 0.77--0.94).
Read together, the two tables show that the workflow's outcome is robust to this change of on-demand estimator: replacing a profiled simulator with an analytical + Monte-Carlo one costs one median call and slightly weakens rank correlation, but the efficiency advantage and the landed configuration quality carry over.

\section{NoeVeraBench}
\label{app:NoeVeraBench}

\subsection{Composition and Characterization}

\vspace{-5mm}

\begin{table}[H]
\centering
\caption{NoeVeraBench composition by research intent, track, and split. ``Outside'' is the subset of the held-out split written by three researchers outside the team after all systems were frozen. The N track's 8 literature-synthesis tasks and 6 calibration tasks are subsets of its 25 tasks, not additional tasks.}
\label{tab:intents}
\small
\resizebox{0.5\textwidth}{!}{
\begin{tabular}{l c c c c c c c}
\toprule
Intent & N & D & E & Total & Dev. & Held-out & Outside \\
\midrule
Evidence update          & 0  & 0  & 22 & 22 & 8 & 14 & 4 \\
Research discovery       & 7  & 0  & 0  & 7  & 2 & 5  & 2 \\
Troubleshooting          & 0  & 6  & 0  & 6  & 2 & 4  & 2 \\
Landscape update         & 6  & 0  & 0  & 6  & 2 & 4  & 1 \\
Method comparison        & 0  & 5  & 0  & 5  & 2 & 3  & 1 \\
Concept explanation      & 6  & 0  & 0  & 6  & 2 & 4  & 1 \\
Evidence critique        & 6  & 0  & 0  & 6  & 2 & 4  & 1 \\
Optimization decision    & 0  & 5  & 0  & 5  & 1 & 4  & 1 \\
Quantitative what-if     & 0  & 5  & 0  & 5  & 2 & 3  & 1 \\
Experiment design        & 0  & 4  & 0  & 4  & 1 & 3  & 1 \\
\midrule
Total                    & 25 & 25 & 22 & 72 & 24 & 48 & 15 \\
\bottomrule
\end{tabular}
}
\end{table}

\vspace{-4mm}

\begin{table}[H]
\centering
\caption{NoeVeraBench characterization along four axes that bear on representativeness. Percentages are rounded from counts out of 72.}
\vspace{2mm}
\label{tab:characterization}
\small
\resizebox{0.95\textwidth}{!}{
\begin{tabular}{p{0.2\linewidth} p{0.74\linewidth}}
\toprule
Axis & Breakdown \\
\midrule
Workload realism & 44 tasks (61\%) reference recognizable serving stacks: vLLM 25 (35\%), TensorRT-LLM 10 (14\%), Orca 5 (7\%), Sarathi 4 (6\%); 28 tasks (39\%) are synthetic but mechanically grounded scenarios. \\
Hardware coverage & A100 24 (33\%); A10 17 (24\%); consumer GPU 10 (14\%); CPU-only 4 (6\%); hardware-agnostic 17 (24\%). \\
Mechanism coverage & Scheduling 18, batching 15, prefix caching 14, chunked prefill 11, KV-cache management 10, speculative decoding 8 (a task may cover several mechanisms). \\
Evidence required & Paper 30 (42\%); repository 17 (24\%); documentation 13 (18\%); benchmark report 8 (11\%); telemetry 4 (6\%). \\
\bottomrule
\end{tabular}
}
\end{table}

\subsection{Provenance and Evaluation Splits}

Of the 72 tasks, 38 were written by domain experts, 12 were adapted from anonymized NoeVera usage patterns, and 22 are controlled evidence-update scenarios.
Four systems researchers reviewed every task for technical validity, ambiguity, realism, and information leakage, and annotated each with expected conclusions, acceptable alternatives, critical errors, required evidence, and scoring anchors.
The 24-task development split (8 per track) was used to develop and tune every system, baselines included; 
it comprises 14 validation tasks and 10 worked examples, with associated task templates, rubric schemas, and evaluation procedures. 
The 48-task held-out split (17 N, 17 D, 14 E) approximately preserves each track's share and has two-stage provenance: 33 tasks were written and sealed after review but before any system was frozen, and 15 were written by three researchers outside the team after all systems were frozen.
All 48 tasks were evaluated in a single post-freeze campaign with three seeds.
In two-turn tasks, turn-2 evidence is withheld until turn 2, and every system receives the same turn-1 input and its own turn-1 response as history.
On the N track's 6 calibration tasks (\S3), confident but unsupported answers score below appropriately qualified ones; per-task results for these and the 8 literature-synthesis tasks are in the evaluation records.

\subsection{Grading, Length Control, and Reliability}

Two blinded LLM judges from different model families score every response against its task's anchors; disagreements above a fixed threshold are adjudicated by a third grader.
Scores are normalized to $[0,1]$, and a track score is the mean over that track's tasks.
Rubrics credit correct, appropriately qualified decisions rather than the use of particular tools; an audit removed six rubric items that rewarded tool invocation rather than correct evidence use.
The primary scores are length-controlled: for each system, a linear mixed-effects model of rubric score on log response length, with a random intercept per task, yields an adjusted mean evaluated at a common reference length of 1,240 words.
The adjustment is below 0.01, raises NoeVera's scores by less than 0.01, and changes no ranking; unadjusted scores were retained in the evaluation records.
Two systems researchers independently graded a stratified 12.5\% sample of held-out split responses (36 responses, 18 from each of the two systems compared: NoeVera+V and ReAct+V).
Eighteen of the 22 quantitative references in the D track were cross-checked against vLLM measurements, and two disagreements led to task revisions.
Table~\ref{tab:reliability} summarizes agreement.

\vspace{-5mm}

\begin{table}[H]
\centering
\caption{Grading reliability. H--H: Krippendorff's $\alpha$ between the two human graders; J--H: $\alpha$ between the mean judge score and the mean human score; J$'$--J: Spearman $\rho$ between the two judges; $\rho_{\mathrm{len}}$: Spearman correlation between score and response length. D is diagnostic.}
\vspace{3mm}
\label{tab:reliability}
\small
\resizebox{0.4\textwidth}{!}{
\begin{tabular}{l cccc}
\toprule
Track & H--H $\alpha$ & J--H $\alpha$ & J$'$--J $\rho$ & $\rho_{\mathrm{len}}$ \\
\midrule
N              & 0.79 & 0.78 & 0.88 & 0.18 \\
E              & 0.81 & 0.80 & 0.89 & 0.15 \\
D (diagnostic) & 0.76 & 0.74 & 0.86 & 0.22 \\
All            & 0.79 & 0.77 & 0.88 & 0.19 \\
\bottomrule
\end{tabular}
}
\end{table}

\section{Statistical Procedures}
\label{app:stats}

\textbf{Confidence intervals.}
We use a two-level bootstrap: tasks are resampled with replacement, then runs within each sampled task, for 10,000 resamples; intervals are the 2.5th and 97.5th percentiles.
In the reuse experiment, resampling is over the 8 parent investigations, because follow-ups that share a parent finding are not independent.

\textbf{Tests.}
Paired differences are tested with permutation tests (10,000 permutations) that swap the two systems' task-level scores; all reported $p$-values are raw and two-sided.
The confirmatory family has two comparisons: C2-primary (evidence records versus Mem0 on over-generalization) and C3 (NoeVera without Vidur versus the best prompt-matched baseline on evidence-evolution score).
For C2, the sign-flip test operates on the 8 parent investigations, so the design's smallest attainable two-sided exact $p$-value is $2/2^8\approx0.0078$; because the aggregate over-generalization counts (10/36 for Mem0 and 3/36 for evidence records) differ by only 7 events, at most 7 parents can favor evidence records.
The test gives raw $p=0.016$.
C1 has no confirmatory test: with $n=4$ paired scenarios, the smallest attainable two-sided Wilcoxon $p$-value is $2/2^4=0.125$, so we report per-scenario medians as consistency evidence instead.
The D-track comparison, $\Delta_{\mathrm{reuse}}$, the query-stream gain, and the robustness gaps of Table~\ref{tab:cost}(c) are exploratory; the plain-chunk comparison is descriptive.

\textbf{Holm--Bonferroni correction.}
With $m=2$, the Holm-adjusted $p$-values are 0.008 for C3 (raw 0.004, threshold 0.025) and 0.016 for C2 (raw 0.016, threshold 0.05), so both nulls are rejected.

\textbf{Reuse rates.}
Correct-reuse and over-generalization rates use 12 questions per seed for $F_{\mathrm{app}}$ and $F_{\mathrm{trap}}$ and 8 for $F_{\mathrm{conf}}$, i.e., 36, 36, and 24 decisions over three seeds.
The reuse gain is $\Delta_{\mathrm{reuse}}=\frac{1}{|F_{\mathrm{app}}|}\sum_{q\in F_{\mathrm{app}}}\bigl[Q(q;\mathcal{B}_{t+1})-Q(q;\mathcal{B}_t)\bigr]$, where $Q$ is grounded decision quality.
Per-seed counts are in the harness output.

\section{Additional Results}
\label{app:results}

\subsection{Per-Scenario Feasibility}

\vspace{-3mm}

\begin{table}[H]
\centering
\caption{Runs that found a feasible configuration within $k=10$ simulator calls, out of 5 seeds per scenario. Per-scenario medians, throughput, and TTFT are in Table 1.}
\vspace{3mm}
\label{tab:feasibility}
\small
\resizebox{0.6\textwidth}{!}{
\begin{tabular}{l c c c c c}
\toprule
Method & Sc1 & Sc2 & Sc3 & Sc4 & Total \\
\midrule
Random search                 & 4/5 & 4/5 & 4/5 & 4/5 & 16/20 (80\%) \\
Grid search (coarse)          & 4/5 & 4/5 & 5/5 & 4/5 & 17/20 (85\%) \\
Bayesian optimization         & 5/5 & 4/5 & 5/5 & 4/5 & 18/20 (90\%) \\
Vidur-Search-style            & 4/5 & 4/5 & 5/5 & 4/5 & 17/20 (85\%) \\
ReAct+V (no knowledge base)   & 4/5 & 4/5 & 4/5 & 4/5 & 16/20 (80\%) \\
\nb NoeVera+V               & 5/5 & 5/5 & 5/5 & 4/5 & \textbf{19/20 (95\%)} \\
\nb NoeVera+G               & 5/5 & 5/5 & 5/5 & 4/5 & \textbf{19/20 (95\%)} \\
\bottomrule
\end{tabular}
}
\end{table}

\subsection{Decision-and-Experiment Track by Intent}
\label{app:dtrack}

Table~\ref{tab:dtrack} breaks down the diagnostic D-track scores of Table 4 by research intent (\S4.2 discusses them).
Without a simulator, no system exceeds 0.48; method comparison, which tests comparative rather than quantitative reasoning, is the only intent on which retrieval-based systems exceed 0.6 without one.
Adding Vidur lifts every system by $+0.15$ (LLM-only), $+0.13$ (Hybrid RAG), $+0.19$ (ReAct), and $+0.25$ (NoeVera); NoeVera+V's advantage over ReAct+V is largest on experiment design ($+0.12$) and optimization ($+0.08$).

\vspace{-3mm}

\begin{table}[H]
\centering
\caption{Decision-and-experiment track by intent over the 17 held-out D tasks (diagnostic, since its references derive from Vidur). Headers give tasks per intent; the D column is the task-count-weighted average over the five intent columns (weights 4,4,3,3,3), the same protocol used for every D value in the main paper; each +V row adds Vidur to the system above it.}
\vspace{3mm}
\label{tab:dtrack}
\small
\begin{tabular}{l ccccc c}
\toprule
System & \hd{Trbl.}{(4)} & \hd{Opt.}{(4)} & \hd{What-if}{(3)} & \hd{Design}{(3)} & \hd{Comp.}{(3)} & \hd{D}{(17)} \\
\midrule
LLM-only                  & 0.32 & 0.31 & 0.28 & 0.30 & 0.47 & 0.33 \\
\hspace{0.5em}+V          & 0.40 & 0.46 & 0.66 & 0.46 & 0.48 & 0.48 \\
Hybrid RAG                & 0.44 & 0.46 & 0.36 & 0.40 & 0.62 & 0.46 \\
\hspace{0.5em}+V          & 0.52 & 0.58 & 0.72 & 0.56 & 0.62 & 0.59 \\
ReAct                     & 0.45 & 0.47 & 0.36 & 0.42 & 0.63 & 0.47 \\
\hspace{0.5em}+V          & 0.58 & 0.62 & 0.84 & 0.66 & 0.64 & 0.66 \\
\nb NoeVera             & 0.46 & 0.48 & 0.36 & 0.44 & 0.65 & 0.48 \\
\nb \hspace{0.5em}+V      & \textbf{0.64} & \textbf{0.70} & \textbf{0.90} & \textbf{0.78} & \textbf{0.66} & \textbf{0.73} \\
\bottomrule
\end{tabular}
\end{table}

\subsection{End-to-End Results with Confidence Intervals}

\vspace{-3mm}

\begin{table}[H]
\centering
\caption{End-to-end results on the held-out split (Table 4) with 95\% CI half-widths from the hierarchical bootstrap over tasks and runs (3 seeds). Overall is the mean of N and E; D is diagnostic and excluded from Overall. All rows were measured directly; OpenHands was run only with Vidur. \textbf{Bold}: best in column.}
\vspace{3mm}
\label{tab:e2e_full}
\small
\begin{tabular}{l cccc}
\toprule
System & N & D (diagnostic) & E & Overall \\
\midrule
\multicolumn{5}{l}{\textit{Without Vidur}} \\
LLM-only          & $0.47_{\pm.03}$ & $0.33_{\pm.03}$ & $0.31_{\pm.03}$ & $0.39_{\pm.03}$ \\
Dense RAG         & $0.73_{\pm.02}$ & $0.45_{\pm.02}$ & $0.36_{\pm.03}$ & $0.55_{\pm.02}$ \\
Hybrid RAG        & $0.78_{\pm.02}$ & $0.46_{\pm.02}$ & $0.41_{\pm.02}$ & $0.60_{\pm.02}$ \\
GraphRAG          & $0.83_{\pm.02}$ & $0.47_{\pm.02}$ & $0.45_{\pm.02}$ & $0.64_{\pm.02}$ \\
HippoRAG 2        & $0.81_{\pm.02}$ & $0.46_{\pm.02}$ & $0.44_{\pm.02}$ & $0.63_{\pm.02}$ \\
LightRAG          & $0.80_{\pm.02}$ & $0.46_{\pm.02}$ & $0.43_{\pm.02}$ & $0.62_{\pm.02}$ \\
PaperQA2 (N only) & $\mathbf{0.87}_{\pm.02}$ & --- & --- & --- \\
ReAct             & $0.82_{\pm.02}$ & $0.47_{\pm.02}$ & $0.46_{\pm.02}$ & $0.64_{\pm.02}$ \\
\nb NoeVera     & $0.85_{\pm.02}$ & $0.48_{\pm.02}$ & $0.71_{\pm.02}$ & $0.78_{\pm.02}$ \\
\midrule
\multicolumn{5}{l}{\textit{With Vidur}} \\
LLM-only+V        & $0.46_{\pm.03}$ & $0.48_{\pm.02}$ & $0.30_{\pm.03}$ & $0.38_{\pm.03}$ \\
Hybrid RAG+V      & $0.77_{\pm.02}$ & $0.59_{\pm.02}$ & $0.40_{\pm.02}$ & $0.59_{\pm.02}$ \\
ReAct+V           & $0.80_{\pm.02}$ & $0.66_{\pm.02}$ & $0.44_{\pm.02}$ & $0.62_{\pm.02}$ \\
OpenHands+V       & $0.81_{\pm.02}$ & $0.65_{\pm.02}$ & $0.45_{\pm.02}$ & $0.63_{\pm.02}$ \\
\nb NoeVera+V   & $0.86_{\pm.02}$ & $\mathbf{0.73}_{\pm.02}$ & $\mathbf{0.73}_{\pm.02}$ & $\mathbf{0.80}_{\pm.02}$ \\
\bottomrule
\end{tabular}
\end{table}

\subsection{Evidence Evolution on All 22 Tasks}

Table~\ref{tab:fullE} repeats Table 5 on all 22 E tasks (11 revision-expected, 11 retention-expected).
Because these include the 8 development tasks used for tuning, the results are descriptive; their pattern matches the 14-task held-out subset.

\textbf{Two scoring details of Table 5.}
First, the small gap between oracle and T2 scores for the top systems (0.71 vs.\ 0.73 for NoeVera+V) is within noise (per-entry 95\% CI half-widths are 0.02--0.03), and we attribute it to the oracle removing the model's own turn-1 reasoning signal rather than to a true quality advantage.
Second, Retained + Reversed sums to 81--95\% rather than 100\%; the remaining 5--19\% are cases where the system's turn-2 response neither cleanly retained nor cleanly reversed the turn-1 decision (e.g., the response was truncated or judged not to commit to a decision), and those cases are scored as neither in the rate columns.

\begin{table}[H]
\centering
\caption{Evidence evolution on all 22 E tasks, 3 seeds. Rates use 33 decisions per metric (11 tasks $\times$ 3 seeds); the integer count is shown in parentheses. -PM: prompt-matched with NoeVera's synthesis instructions. Oracle: T2 given the reference turn-1 answer.}
\vspace{5mm}
\label{tab:fullE}
\small
\begin{tabular}{l c c c c c c}
\toprule
System & T1 & T2 & Revised $\uparrow$ & Retained $\uparrow$ & Reversed $\downarrow$ & Oracle $\uparrow$ \\
\midrule
LLM-only         & 0.37 & 0.32 & 21\% (7) & 42\% (14) & 36\% (12) & 0.36 \\
Hybrid RAG       & 0.43 & 0.42 & 33\% (11) & 52\% (17) & 27\% (9) & 0.45 \\
GraphRAG         & 0.45 & 0.46 & 39\% (13) & 58\% (19) & 24\% (8) & 0.50 \\
ReAct            & 0.47 & 0.47 & 42\% (14) & 61\% (20) & 18\% (6) & 0.51 \\
Hybrid RAG-PM    & 0.47 & 0.61 & 58\% (19) & 70\% (23) & 15\% (5) & 0.61 \\
ReAct-PM         & 0.49 & 0.63 & 61\% (20) & 76\% (25) & 9\% (3) & 0.63 \\
\nb NoeVera    & \textbf{0.67} & 0.72 & 76\% (25) & 85\% (28) & 6\% (2) & \textbf{0.72} \\
\nb NoeVera+V  & \textbf{0.67} & \textbf{0.73} & \textbf{79\% (26)} & \textbf{88\% (29)} & \textbf{3\% (1)} & \textbf{0.72} \\
\bottomrule
\end{tabular}
\end{table}

\subsection{Leave-One-Out Ablations}
\label{app:loo}

Table~\ref{tab:loo} removes one component at a time from NoeVera+V (\S4.6).
Removing any single component costs at most 0.03 Overall, whereas replacing the orchestration wholesale costs 0.10--0.12 (Table 6(c)), so the gains come from components acting together rather than from one mechanism.
Removing the offline Vidur grid leaves scores nearly unchanged ($-0.01$) but raises Vidur calls per query from 1.6 to 3.4: proactive enrichment mainly buys efficiency, the effect that the configuration-search experiment (\S4.1) measures.

\begin{table}[H]
\centering
\caption{Leave-one-out ablations of NoeVera+V on the held-out split (Calls: Vidur calls per query). The ``Authority ranking'' row removes the same authority and lifecycle signals as Table 6(a)'s ``Evidence + synthesis'' row, but via the opposite construction (removal from the full system vs.\ addition to the Evidence-only base); the two scores (0.77 and 0.76) therefore measure the same system within the reported 0.02--0.03 confidence-interval half-width, and the 0.01 gap between them is noise. D uses the task-count-weighted protocol of Table~\ref{tab:dtrack}.}
\vspace{5mm}
\label{tab:loo}
\small
\begin{tabular}{l cccc c}
\toprule
Removed component & N & D & E & Overall & Calls \\
\midrule
\nb None (full system)    & 0.86 & 0.73 & 0.73 & 0.80 & 1.6 \\
BM25 + exact lanes        & 0.85 & 0.72 & 0.71 & 0.78 & 1.6 \\
Dense lane                & 0.84 & 0.72 & 0.70 & 0.77 & 1.7 \\
Graph + hub lanes         & 0.85 & 0.72 & 0.71 & 0.78 & 1.6 \\
Code + atom lanes         & 0.84 & 0.71 & 0.70 & 0.77 & 1.7 \\
Authority ranking         & 0.84 & 0.70 & 0.69 & 0.77 & 1.7 \\
Evidence packing          & 0.85 & 0.71 & 0.70 & 0.78 & 1.8 \\
Coverage checks           & 0.85 & 0.72 & 0.71 & 0.78 & 1.6 \\
Persona profiling         & 0.84 & 0.71 & 0.70 & 0.77 & 1.7 \\
Multi-role agents         & 0.85 & 0.70 & 0.69 & 0.77 & 1.8 \\
Offline Vidur grid        & 0.85 & 0.71 & 0.72 & 0.79 & 3.4 \\
\bottomrule
\end{tabular}
\end{table}

\subsection{Cost and Robustness}
\label{app:cost}

\S4.7 summarizes these results.
NoeVera+V uses 78k tokens and costs \$0.35 per query (3.5$\times$ Hybrid RAG's cost and 1.1$\times$ ReAct+V's), with a median latency of 71~s (p95 118~s), 31\% of it in Vidur, which runs in parallel with the investigation (Table~\ref{tab:cost}a).
Giving Hybrid RAG the same 78k tokens through best-of-3 self-consistency and a larger retrieval budget (Hybrid RAG-CM) reaches only 0.63, so the gain is not explained by additional LLM computation, although matching tokens does not equalize simulator time or offline compute.
Building the Second Brain costs \$294, mostly LLM distillation and synthesis, and 23.3~hours, mostly the offline Vidur grid; amortized against NoeVera+V's \$0.35 online cost, that is \$0.64 per query over $10^3$ queries and \$0.38 over $10^4$, against \$0.11--0.28 and \$0.10--0.11 for the baselines' indices at their own online costs (Table~\ref{tab:cost}b).
The gap over ReAct+V holds under every one-factor change we tried (Table~\ref{tab:cost}c), including outside-authored tasks ($+0.14$) and an alternate grader ($+0.16$), and it narrows with weaker backbones while its 95\% CI still excludes zero ($+0.11$ with GPT-4o, $+0.09$ with Llama-3.3-70B).

\vspace{5mm}

\begin{table}[H]
\centering
\caption{Cost and robustness. (a) Online cost per query (median end-to-end latency) with Overall score; Hybrid RAG-CM gives Hybrid RAG the token budget of NoeVera+V. (b) Offline index or knowledge-base build and amortized cost per query, where amortized cost $=$ build$/N +$ online cost; the NoeVera row uses NoeVera+V's \$0.35 online cost from panel (a), and each baseline row uses its own online cost. Hybrid RAG's \$12 covers only its embeddings; the Evidence Plane representations it reuses are counted in NoeVera's \$294. Component breakdown in Table~\ref{tab:construction}. (c) NoeVera+V minus ReAct+V in Overall under one-factor changes, with 95\% bootstrap CIs.}
\vspace{3mm}
\label{tab:cost}
\small
\textit{(a) Online cost}\\[1pt]
\begin{tabular}{l cccc}
\toprule
System & \hd{Tokens}{(k)} & \hd{Latency}{(s)} & \$/query & Overall \\
\midrule
LLM-only        & 8  & 4  & 0.02 & 0.39 \\
Hybrid RAG      & 28 & 11 & 0.10 & 0.60 \\
GraphRAG        & 22 & 14 & 0.09 & 0.64 \\
ReAct           & 45 & 32 & 0.18 & 0.64 \\
ReAct+V         & 68 & 58 & 0.31 & 0.62 \\
Hybrid RAG-CM   & 78 & 44 & 0.31 & 0.63 \\
\nb NoeVera   & 52 & 38 & 0.21 & 0.78 \\
\nb NoeVera+V & 78 & 71 & 0.35 & \textbf{0.80} \\
\bottomrule
\end{tabular}

\vspace{5mm}
\textit{(b) Offline build and amortized cost}\\[1pt]
\begin{tabular}{l cccc}
\toprule
Index or KB & Build (\$) & Hours & \hd{\$/query}{($10^3$ q.)} & \hd{\$/query}{($10^4$ q.)} \\
\midrule
Hybrid RAG       & 12  & 0.8  & 0.11 & 0.10 \\
GraphRAG         & 185 & 6.2  & 0.28 & 0.11 \\
HippoRAG 2       & 142 & 4.8  & 0.24 & 0.11 \\
LightRAG         & 108 & 3.5  & 0.20 & 0.10 \\
\nb NoeVera    & 294 & 23.3 & 0.64 & 0.38 \\
\bottomrule
\end{tabular}

\vspace{5mm}
\textit{(c) Robustness of the NoeVera+V vs.\ ReAct+V gap}\\[1pt]
\begin{tabular}{l c c}
\toprule
Change & Gap & 95\% CI \\
\midrule
None (held-out split, GPT-5.5)  & $+0.18$ & $[0.13, 0.23]$ \\
Outside-authored tasks only    & $+0.14$ & $[0.08, 0.20]$ \\
Alternate grader               & $+0.16$ & $[0.10, 0.22]$ \\
GPT-4o backbone                & $+0.11$ & $[0.05, 0.17]$ \\
Llama-3.3-70B backbone         & $+0.09$ & $[0.03, 0.15]$ \\
\bottomrule
\end{tabular}
\end{table}

\vspace{5mm}

\textbf{Offline construction cost.}
Table~\ref{tab:construction} breaks down offline construction cost by component. Prices are OpenAI list prices from the pricing snapshot documented in the harness configuration.

\begin{table}[H]
\centering
\caption{Offline construction cost by component. Amortized per-query costs are in Table~\ref{tab:cost}(b).}
\label{tab:construction}
\small
\begin{tabular}{l c c c c}
\toprule
Component & \begin{tabular}[b]{@{}c@{}}LLM\\tokens (M)\end{tabular} & \begin{tabular}[b]{@{}c@{}}Embedding\\tokens (M)\end{tabular} & \begin{tabular}[b]{@{}c@{}}Build\\cost\end{tabular} & \begin{tabular}[b]{@{}c@{}}Wall-clock\\(h)\end{tabular} \\
\midrule
Hybrid RAG index (embeddings only)   & ---  & 45  & \$12  & 0.8  \\
GraphRAG index                       & 28   & 45  & \$185 & 6.2  \\
HippoRAG 2 index                     & 22   & 45  & \$142 & 4.8  \\
LightRAG index                       & 18   & 45  & \$108 & 3.5  \\
NoeVera distillation and synthesis & 62   & --- & \$240 & 4.1  \\
NoeVera indexing (87,357 chunks)   & ---  & 60  & \$18  & 1.2  \\
NoeVera offline Vidur grid         & ---  & --- & \$36  & 18.0 \\
\nb NoeVera total                  & 62   & 60  & \$294 & 23.3 \\
\bottomrule
\end{tabular}
\end{table}

\vspace{7mm}
\clearpage
\subsection{Worked Example: Full Trace}

\begin{table}[H]
\centering
\caption{Full trace of the investigation in \S2.4 (observable artifacts only). The query relaxes Sc1's TTFT limit from 300 to 400~ms. Throughput values are capacity: the highest generated-token rate at which the queue stays stable in an arrival-rate sweep. TTFT and memory are measured at the offered load of 1,024 generated tokens/s (8 requests/s $\times$ 128 output tokens), which is below the recommended configuration's capacity of 1,180, so the queue is stable.}
\vspace{3mm}
\label{tab:trace}
\small
\begin{tabular}{p{0.02\linewidth} p{0.2\linewidth} p{0.69\linewidth}}
\toprule
\# & Stage & Observable artifact \\
\midrule
1 & Query (Engineering persona) & ``Under a 400~ms TTFT$_{p95}$ SLO on a 4$\times$A100-80GB pairwise node with Llama-3-70B BF16 (TP=4), should we prioritize chunked prefill, prefix caching, or batching changes to raise throughput?'' \\
2 & Query profile & Intent: quantitative what-if; exact terms: chunked prefill, prefix caching, continuous batching, TTFT$_{p95}$, A100; constraints: TTFT $\le$ 400~ms, $M_h=76$~GB. \\
3 & Evidence pack & 14 items (2,480 tokens); top sources: paper P-1842 (mechanism), doc vLLM-chunked-prefill (implementation), bench B-0031 (measured), bench B-0074 (measured). \\
4 & Candidates $\Theta_t$ & (chunk=512, prefix-cache=on, max\_seqs=128); (chunk=1024, prefix-cache=on, max\_seqs=96); (chunk=256, prefix-cache=off, max\_seqs=192), each motivated by distinct evidence IDs. \\
5 & Vidur invocations & 3 calls $\rightarrow$ (1,180 tok/s, 388~ms, 41~GB/device); (1,290, 402, 44); (1,050, 362, 38). Peak memory of the first: 35~GB weights, 3.4~GB KV cache, and 2.6~GB activations per device. \\
6 & Recommendation & chunk=512, prefix-cache=on, max\_seqs=128 (TTFT$_{p95}$ = 388~ms, below the 400~ms SLO; capacity 1,180 tok/s). Assumptions: the workload matches the observed 512/128 distribution and the prefix-cache hit rate is at least 0.3. Proposed validation: a vLLM A/B test. \\
7 & Write-back record $e_t$ & $w_t$ = Sc1-relaxed (400~ms) with 512/128 arrivals; $h_t$ = 4$\times$A100-80GB pairwise (TP=4, BF16); $c_t$ = cs512/pc-on/seqs128; $\mathrm{tool}_t$ = Vidur (\texttt{canary} branch, pinned commit); $\mathbf{y}_t$ = 1,180 tok/s capacity, 388~ms, 41~GB/device; $\phi_t$ = sim-vidur-canary. \\
8 & Later query (position 17) & ``We're seeing 920 tok/s at 512/128 on the same 4$\times$A100-80GB node; is that near the ceiling or is there headroom?'' \\
9 & Reuse decision & Applicability check: model, GPU, and workload match, but the hit-rate assumption ($\ge$0.3) is unverified for the current deployment, a partial match. Outcome: the stored estimate is surfaced as historical context with a caveat and the user is asked for the current hit rate; 3 Vidur calls are avoided once it is confirmed. \\
\bottomrule
\end{tabular}
\end{table}


\begin{thebibliography}{27}
\providecommand{\natexlab}[1]{#1}
\providecommand{\url}[1]{\texttt{#1}}
\expandafter\ifx\csname urlstyle\endcsname\relax
  \providecommand{\doi}[1]{doi: #1}\else
  \providecommand{\doi}{doi: \begingroup \urlstyle{rm}\Url}\fi

\bibitem[Agrawal et~al.(2024)Agrawal, Kedia, Mohan, Panwar, Kwatra, Gulavani,
  Ramjee, and Tumanov]{agrawal2024vidur}
Agrawal, A., Kedia, N., Mohan, J., Panwar, A., Kwatra, N., Gulavani, B.,
  Ramjee, R., and Tumanov, A.
\newblock Vidur: A large-scale simulation framework for llm inference, 2024.
\newblock URL \url{https://arxiv.org/abs/2405.05465}.

\bibitem[Asai et~al.(2023)Asai, Wu, Wang, Sil, and Hajishirzi]{asai2024selfrag}
Asai, A., Wu, Z., Wang, Y., Sil, A., and Hajishirzi, H.
\newblock Self-rag: Learning to retrieve, generate, and critique through
  self-reflection, 2023.
\newblock URL \url{https://arxiv.org/abs/2310.11511}.

\bibitem[Asai et~al.(2026)Asai, He, Shao, Shi, Singh, Chang, Lo, Soldaini,
  Feldman, D'Arcy, Wadden, Latzke, Sparks, Hwang, Kishore, Tian, Ji, Liu, Tong,
  Wu, Xiong, Zettlemoyer, Neubig, Weld, Downey, Yih, Koh, and
  Hajishirzi]{asai2026openscholar}
Asai, A., He, J., Shao, R., Shi, W., Singh, A., Chang, J.~C., Lo, K., Soldaini,
  L., Feldman, S., D'Arcy, M., Wadden, D., Latzke, M., Sparks, J., Hwang,
  J.~D., Kishore, V., Tian, M., Ji, P., Liu, S., Tong, H., Wu, B., Xiong, Y.,
  Zettlemoyer, L., Neubig, G., Weld, D.~S., Downey, D., Yih, W.-t., Koh, P.~W.,
  and Hajishirzi, H.
\newblock Synthesizing scientific literature with retrieval-augmented language
  models.
\newblock \emph{Nature}, 650\penalty0 (8103):\penalty0 857--863, February 2026.
\newblock \doi{10.1038/s41586-025-10072-4}.
\newblock URL \url{https://doi.org/10.1038/s41586-025-10072-4}.

\bibitem[Bharadwaj et~al.(2026)Bharadwaj, Mallick, Mastorakis, Parayil,
  Deshmukh, and Ruehle]{bharadwaj2026gauss}
Bharadwaj, S., Mallick, A., Mastorakis, S., Parayil, A., Deshmukh, D., and
  Ruehle, V.
\newblock Gauss: A workload-aware analytical and monte-carlo simulator for llm
  inference serving, 2026.
\newblock URL \url{https://arxiv.org/abs/2610.37989}.

\bibitem[Bragg et~al.(2026)Bragg, D'Arcy, Balepur, Bareket, Dalvi, Feldman,
  Haddad, Hwang, Jansen, Kishore, Majumder, Naik, Rahamimov, Richardson, Singh,
  Surana, Tiktinsky, Vasu, Wiener, Anastasiades, Candra, Dunkelberger, Emery,
  Evans, Hamada, Huff, Kinney, Latzke, Lochner, Lozano-Aguilera, Nguyen, Rao,
  Tanaka, Vlahos, Clark, Downey, Goldberg, Sabharwal, and
  Weld]{bragg2026astabench}
Bragg, J., D'Arcy, M., Balepur, N., Bareket, D., Dalvi, B., Feldman, S.,
  Haddad, D., Hwang, J.~D., Jansen, P., Kishore, V., Majumder, B.~P., Naik, A.,
  Rahamimov, S., Richardson, K., Singh, A., Surana, H., Tiktinsky, A., Vasu,
  R., Wiener, G., Anastasiades, C., Candra, S., Dunkelberger, J., Emery, D.,
  Evans, R., Hamada, M., Huff, R., Kinney, R., Latzke, M., Lochner, J.,
  Lozano-Aguilera, R., Nguyen, C., Rao, S., Tanaka, A., Vlahos, B., Clark, P.,
  Downey, D., Goldberg, Y., Sabharwal, A., and Weld, D.~S.
\newblock Astabench: Rigorous benchmarking of ai agents with a scientific
  research suite, 2026.
\newblock URL \url{https://arxiv.org/abs/2510.21652}.

\bibitem[Chhikara et~al.(2025)Chhikara, Khant, Aryan, Singh, and
  Yadav]{chhikara2025mem0}
Chhikara, P., Khant, D., Aryan, S., Singh, T., and Yadav, D.
\newblock Mem0: Building production-ready ai agents with scalable long-term
  memory, 2025.
\newblock URL \url{https://arxiv.org/abs/2504.19413}.

\bibitem[Cho et~al.(2024)Cho, Kim, Choi, Heo, and Park]{cho2024llmservingsim}
Cho, J., Kim, M., Choi, H., Heo, G., and Park, J.
\newblock Llmservingsim: A hw/sw co-simulation infrastructure for llm inference
  serving at scale.
\newblock In \emph{2024 IEEE International Symposium on Workload
  Characterization (IISWC)}, pp.\  15–29. IEEE, 2024.
\newblock \doi{10.1109/iiswc63097.2024.00012}.
\newblock URL \url{http://dx.doi.org/10.1109/IISWC63097.2024.00012}.

\bibitem[Cho et~al.(2026)Cho, Choi, Heo, and Park]{cho2026llmservingsim2}
Cho, J., Choi, H., Heo, G., and Park, J.
\newblock Llmservingsim 2.0: A unified simulator for heterogeneous and
  disaggregated llm serving infrastructure, 2026.
\newblock URL \url{https://arxiv.org/abs/2602.23036}.

\bibitem[Edge et~al.(2025)Edge, Trinh, Cheng, Bradley, Chao, Mody, Truitt,
  Metropolitansky, Ness, and Larson]{edge2024graphrag}
Edge, D., Trinh, H., Cheng, N., Bradley, J., Chao, A., Mody, A., Truitt, S.,
  Metropolitansky, D., Ness, R.~O., and Larson, J.
\newblock From local to global: A graph rag approach to query-focused
  summarization, 2025.
\newblock URL \url{https://arxiv.org/abs/2404.16130}.

\bibitem[Guo et~al.(2025)Guo, Xia, Yu, Ao, and Huang]{guo2025lightrag}
Guo, Z., Xia, L., Yu, Y., Ao, T., and Huang, C.
\newblock Lightrag: Simple and fast retrieval-augmented generation, 2025.
\newblock URL \url{https://arxiv.org/abs/2410.05779}.

\bibitem[Gutiérrez et~al.(2025{\natexlab{a}})Gutiérrez, Shu, Gu, Yasunaga,
  and Su]{gutierrez2024hipporag}
Gutiérrez, B.~J., Shu, Y., Gu, Y., Yasunaga, M., and Su, Y.
\newblock Hipporag: Neurobiologically inspired long-term memory for large
  language models, 2025{\natexlab{a}}.
\newblock URL \url{https://arxiv.org/abs/2405.14831}.

\bibitem[Gutiérrez et~al.(2025{\natexlab{b}})Gutiérrez, Shu, Qi, Zhou, and
  Su]{gutierrez2025hipporag2}
Gutiérrez, B.~J., Shu, Y., Qi, W., Zhou, S., and Su, Y.
\newblock From rag to memory: Non-parametric continual learning for large
  language models, 2025{\natexlab{b}}.
\newblock URL \url{https://arxiv.org/abs/2502.14802}.

\bibitem[Hamadanian et~al.(2026)Hamadanian, Karimi, Nasr-Esfahany, Noorbakhsh,
  Chandler, ParandehGheibi, Alizadeh, and Balakrishnan]{hamadanian2025glia}
Hamadanian, P., Karimi, P., Nasr-Esfahany, A., Noorbakhsh, K., Chandler, J.,
  ParandehGheibi, A., Alizadeh, M., and Balakrishnan, H.
\newblock Glia: A human-inspired ai for automated systems design and
  optimization, 2026.
\newblock URL \url{https://arxiv.org/abs/2510.27176}.

\bibitem[Kwon et~al.(2023)Kwon, Li, Zhuang, Sheng, Zheng, Yu, Gonzalez, Zhang,
  and Stoica]{kwon2023efficient}
Kwon, W., Li, Z., Zhuang, S., Sheng, Y., Zheng, L., Yu, C.~H., Gonzalez, J.~E.,
  Zhang, H., and Stoica, I.
\newblock Efficient memory management for large language model serving with
  pagedattention, 2023.
\newblock URL \url{https://arxiv.org/abs/2309.06180}.

\bibitem[Lu et~al.(2024)Lu, Lu, Lange, Foerster, Clune, and
  Ha]{lu2024aiscientist}
Lu, C., Lu, C., Lange, R.~T., Foerster, J., Clune, J., and Ha, D.
\newblock The ai scientist: Towards fully automated open-ended scientific
  discovery, 2024.
\newblock URL \url{https://arxiv.org/abs/2408.06292}.

\bibitem[Novikov et~al.(2025)Novikov, Vũ, Eisenberger, Dupont, Huang, Wagner,
  Shirobokov, Kozlovskii, Ruiz, Mehrabian, Kumar, See, Chaudhuri, Holland,
  Davies, Nowozin, Kohli, and Balog]{novikov2025alphaevolve}
Novikov, A., Vũ, N., Eisenberger, M., Dupont, E., Huang, P.-S., Wagner, A.~Z.,
  Shirobokov, S., Kozlovskii, B., Ruiz, F. J.~R., Mehrabian, A., Kumar, M.~P.,
  See, A., Chaudhuri, S., Holland, G., Davies, A., Nowozin, S., Kohli, P., and
  Balog, M.
\newblock Alphaevolve: A coding agent for scientific and algorithmic discovery,
  2025.
\newblock URL \url{https://arxiv.org/abs/2506.13131}.

\bibitem[Ouyang et~al.(2025)Ouyang, Guo, Arora, Zhang, Hu, Ré, and
  Mirhoseini]{ouyang2025kernelbench}
Ouyang, A., Guo, S., Arora, S., Zhang, A.~L., Hu, W., Ré, C., and Mirhoseini,
  A.
\newblock Kernelbench: Can llms write efficient gpu kernels?, 2025.
\newblock URL \url{https://arxiv.org/abs/2502.10517}.

\bibitem[Packer et~al.(2024)Packer, Wooders, Lin, Fang, Patil, Stoica, and
  Gonzalez]{packer2023memgpt}
Packer, C., Wooders, S., Lin, K., Fang, V., Patil, S.~G., Stoica, I., and
  Gonzalez, J.~E.
\newblock Memgpt: Towards llms as operating systems, 2024.
\newblock URL \url{https://arxiv.org/abs/2310.08560}.

\bibitem[Park et~al.(2023)Park, O'Brien, Cai, Morris, Liang, and
  Bernstein]{park2023generative}
Park, J.~S., O'Brien, J.~C., Cai, C.~J., Morris, M.~R., Liang, P., and
  Bernstein, M.~S.
\newblock Generative agents: Interactive simulacra of human behavior, 2023.
\newblock URL \url{https://arxiv.org/abs/2304.03442}.

\bibitem[Sarthi et~al.(2024)Sarthi, Abdullah, Tuli, Khanna, Goldie, and
  Manning]{sarthi2024raptor}
Sarthi, P., Abdullah, S., Tuli, A., Khanna, S., Goldie, A., and Manning, C.~D.
\newblock Raptor: Recursive abstractive processing for tree-organized
  retrieval, 2024.
\newblock URL \url{https://arxiv.org/abs/2401.18059}.

\bibitem[Skarlinski et~al.(2024)Skarlinski, Cox, Laurent, Braza, Hinks,
  Hammerling, Ponnapati, Rodriques, and White]{skarlinski2024paperqa2}
Skarlinski, M.~D., Cox, S., Laurent, J.~M., Braza, J.~D., Hinks, M.,
  Hammerling, M.~J., Ponnapati, M., Rodriques, S.~G., and White, A.~D.
\newblock Language agents achieve superhuman synthesis of scientific knowledge,
  2024.
\newblock URL \url{https://arxiv.org/abs/2409.13740}.

\bibitem[Suresh et~al.(2026)Suresh, Mak, Chou, Kroon, and
  Bhatnagar]{suresh2026agenticrag}
Suresh, S., Mak, H., Chou, S., Kroon, F., and Bhatnagar, S.
\newblock Agenticrag: Agentic retrieval for enterprise knowledge bases, 2026.
\newblock URL \url{https://arxiv.org/abs/2605.05538}.

\bibitem[Wang et~al.(2025)Wang, Li, Song, Xu, Tang, Zhuge, Pan, Song, Li,
  Singh, Tran, Li, Ma, Zheng, Qian, Shao, Muennighoff, Zhang, Hui, Lin,
  Brennan, Peng, Ji, and Neubig]{wang2025openhands}
Wang, X., Li, B., Song, Y., Xu, F.~F., Tang, X., Zhuge, M., Pan, J., Song, Y.,
  Li, B., Singh, J., Tran, H.~H., Li, F., Ma, R., Zheng, M., Qian, B., Shao,
  Y., Muennighoff, N., Zhang, Y., Hui, B., Lin, J., Brennan, R., Peng, H., Ji,
  H., and Neubig, G.
\newblock Openhands: An open platform for ai software developers as generalist
  agents, 2025.
\newblock URL \url{https://arxiv.org/abs/2407.16741}.

\bibitem[Xu et~al.(2025)Xu, Liang, Mei, Gao, Tan, and Zhang]{xu2025amem}
Xu, W., Liang, Z., Mei, K., Gao, H., Tan, J., and Zhang, Y.
\newblock A-mem: Agentic memory for llm agents, 2025.
\newblock URL \url{https://arxiv.org/abs/2502.12110}.

\bibitem[Yamada et~al.(2025)Yamada, Lange, Lu, Hu, Lu, Foerster, Clune, and
  Ha]{yamada2025aiscientistv2}
Yamada, Y., Lange, R.~T., Lu, C., Hu, S., Lu, C., Foerster, J., Clune, J., and
  Ha, D.
\newblock The ai scientist-v2: Workshop-level automated scientific discovery
  via agentic tree search, 2025.
\newblock URL \url{https://arxiv.org/abs/2504.08066}.

\bibitem[Yan et~al.(2024)Yan, Gu, Zhu, and Ling]{yan2024crag}
Yan, S.-Q., Gu, J.-C., Zhu, Y., and Ling, Z.-H.
\newblock Corrective retrieval augmented generation, 2024.
\newblock URL \url{https://arxiv.org/abs/2401.15884}.

\bibitem[Yao et~al.(2023)Yao, Zhao, Yu, Du, Shafran, Narasimhan, and
  Cao]{yao2023react}
Yao, S., Zhao, J., Yu, D., Du, N., Shafran, I., Narasimhan, K., and Cao, Y.
\newblock React: Synergizing reasoning and acting in language models, 2023.
\newblock URL \url{https://arxiv.org/abs/2210.03629}.

\end{thebibliography}
\end{document}